%% file: neurips_2026.tex
\documentclass{article}

\usepackage[numbers,sort&compress]{natbib}

\usepackage[preprint, nonatbib]{neurips_2026}

\usepackage[utf8]{inputenc} 
\usepackage[T1]{fontenc}    
\usepackage{url}            
\usepackage{booktabs}       
\usepackage{amsfonts}       
\usepackage{nicefrac}       
\usepackage{microtype}      
\usepackage{xcolor}         
\usepackage[most]{tcolorbox}
\usepackage{caption}

\usepackage{float}
\usepackage{tabularx}          
\usepackage{makecell}          
\usepackage{tikz}
\usepackage{tcolorbox}
\usepackage{xcolor}
\definecolor{lightblue}{rgb}{0.2, 0.4, 0.8}
\definecolor{darkerblue}{HTML}{2244AA}
\definecolor{lightred}{HTML}{CC3366}
\usepackage[colorlinks=true, citecolor=darkerblue, urlcolor=lightred]{hyperref}
\usepackage{amsmath}

\newcolumntype{Z}{>{\centering\arraybackslash}X}

\usepackage{fontawesome5}

\newcommand{\iconhref}[3]{%
  \href{#2}{\faIcon{#1}\,\,#3}%
}

\usepackage{enumitem}
\setlist[itemize]{itemsep=1pt, topsep=0pt, leftmargin=*}
\setlist[enumerate]{itemsep=1pt, topsep=0pt, leftmargin=*}

\title{Benchmarking In-context Experiential Learning \\
    Through Repeated Product Recommendations}

\author{%
  Gilbert Yang\thanks{Equal contribution.} \\
  Decision, Risk, and Operations Division\\
  Columbia Business School\\
  \texttt{gy2322@columbia.edu} \\
  \And
  Yaqin Chen\footnotemark[1] \\
  School of Mathematics (Zhuhai)\\
  Sun Yat-sen University\\
  \texttt{chenyq077@gmail.com} \\
  \And
  Thomson Yen\footnotemark[1] \\
  Decision, Risk, and Operations Division\\
  Columbia Business School\\
  \texttt{ty2531@columbia.edu} \\
  \And
  Hongseok Namkoong \\
  Decision, Risk, and Operations Division\\
  Columbia Business School\\
  \texttt{hongseok.namkoong@columbia.edu} \\
}

\begin{document}

\maketitle

\begin{abstract}
  \input{./sections/00_abstract}
\end{abstract}

\section{Introduction}
\label{sec:intro}
\input{./sections/01_introduction}

\section{Recommendation As A Testbed for Experiential Learning}
\label{sec:problem_setup}
\input{./sections/problem_setup}

\section{BELA: A Benchmark for Experiential Learning and Active Exploration}
\label{sec:benchmark_setup}
\input{./sections/benchmark_setup}

\section{Experiments}
\label{sec:experiments}
\input{./sections/experiments}

\section{Related Work}
\label{sec:related_works}
\input{./sections/related_works}

\section{Conclusion}
\input{./sections/conclusion}

\bibliographystyle{unsrtnat}
\bibliography{bib2} 

\appendix

\input{./sections/appendix}


\end{document}

%% file: sections/00_abstract.tex
To navigate ever-shifting real-world environments, agents must grapple with incomplete knowledge and adapt their strategies through \emph{experience}. 
However, current evaluations of LLM-based agents largely overlook this capability.
Crucially, we stress not just the ability to contend with uncertainty within a task (episode), as episode-specific information is progressively revealed across turns, but also the refinement of such adaptive ability across similar episodes, as agents accumulate experiences and infer their shared latent structure. 
Repeated product recommendation offers a natural setting to isolate this need for \textit{experiential learning}:
within each interaction, an intelligent recommender must elicit unknown customer preferences through questions; across multiple interactions, it should tailor its questioning strategy to the observed customer and product distributions.
We instantiate the \textbf{B}enchmark for \textbf{E}xperiential \textbf{L}earning and \textbf{A}ctive exploration (\textbf{BELA}) by combining (1) a rich catalog of real-world products from Amazon, (2) a diverse collection of synthetic customer personas aimed to capture heterogeneous latent preferences, and (3) an LLM-based customer simulator framework that emulate preference-revealing interactions. 
Rather than aiming to faithfully replicate real consumer behavior, BELA provides a controlled and scalable testbed for whether agents can exploit consistent latent preferences across episodes. 
Benchmarking current models reveals that they can learn across turns, but struggle to improve across episodes. 
This underscore the need for frontier models to advance in experiential learning capabilities.

\vspace{0.2cm}
\begin{center}
\setlength{\tabcolsep}{1.2em}
\begin{tabular}{@{}ll@{}}
\iconhref{globe}{https://www.experiential-learning-benchmark.com}{Website} &
\iconhref{github}{https://github.com/namkoong-lab/interactive-benchmark}{Code}
\end{tabular}
\end{center}

%% file: sections/01_introduction.tex

The ability to acquire and to learn from experience is a hallmark of intelligence.
Real-world environments involve uncertainty arising from unobserved information, and intelligent agents must deliberately act to reduce unknowns and to learn from acquired observations.
Recently, several benchmarks have introduced tasks with unknowns that agents should strategically resolve in a multi-turn task \citep{li2024mediq, barres2025tau2,  zhou2025passiveactivereasoninglarge, deoliveira2025infoquestevaluatingmultiturndialogue, ramrakhya2025groundingmultimodalllmsembodied, liu-etal-2025-interactive, fang2026benchmarkingmultiturnmedicaldiagnosis, luo2025clarifymtbenchbenchmarkingimprovingmultiturn}. 
In addition to learning within a task, real-world agents have the opportunity to refine their strategy from inferring the shared uncertainty structure in observed tasks.
However, measuring and advancing this ability has received far less attention.


To motivate this ability, consider a sales agent recommending products to a stream of incoming unknown customers. 
Early on, the agent must probe broadly to identify each customer's taste. 
But after repeated recommendations, the agent should recognize that its user base concentrates on a few dominant preference clusters, and that a small number of well-chosen questions reliably disambiguate among them. 
An agent deployed on a niche marketplace should instead learn the opposite: as preferences are idiosyncratic, the same broad questions are uninformative. 
The structure of the latent preference distribution dictates how the agent should act under uncertainty, and since this structure is rarely known a priori, agents must learn to infer and exploit it from experience. 
Following~\citet{silver2025welcome}, we refer to this ability as \emph{experiential learning}.
Without it, large language model (LLM) agents cannot tailor their information-gathering to the observed task distribution at deployment. 
We stress the need to examine this capability \textit{in-context}, since the optimal strategy depends on distribution of tasks revealed at test time. 

We construct the \textbf{B}enchmark for \textbf{E}xperiential \textbf{L}earning and \textbf{A}ctive exploration (\textbf{BELA})  that isolates this capacity for in-context experiential learning in product recommendation setting.
BELA consists of two main modules: (1) a \textit{choice set} module with numerous sets of similar products that a customer might choose from, 
and (2) a \textit{customer} module that admits varying latent preferences and provide interactive responses to recommenders. 
The customer module is designed for learnability rather than realism: it exposes consistent latent structure that the recommender can learn from interactions.
By combining real amazon products \citep{hou2024_amazonproducts} and diverse synthetic LLM persona \citep{li2025llmgeneratedpersonapromise}, BELA supports up to 2 billion unique choice set–persona pairs. 
Each pairing constitutes a single task with natural uncertainty in the customer's general preferences and their reactions to the available products. 
By exposing agents on a stream of such pairings, combinatorially composable from this space of 2 billion pairs, BELA offers an uniquely scalable environments for measuring agent's experiential learning capability under recommendation uncertainty. 
See Figure \ref{fig:multi_episode_examples_unified} for an example of our setting.

Our main contributions are as follows:
\begin{itemize}
\item We formulate in-context experiential learning as a centerpiece to intelligence. Agents must strategically resolve uncertainty within a task (across turns), and refine their strategies by leveraging past task experiences (across episodes). See Figure~\ref{fig:setup}. 
\item To measure in-context experiential learning capability, we develop a \textbf{B}enchmark for \textbf{E}xperiential \textbf{L}earning and \textbf{A}ctive exploration (\textbf{BELA}), consisting of diverse and scalable product categories with 71K products and 2K choice sets. 
Coupled with 1M scalably generated personas from \citet{li2025llmgeneratedpersonapromise}, the benchmark supports up to 2B multi-turn product recommendation environments on which one can combinatorially scale experiential learning environments.

\item We observe that while state-of-the-art models exhibit learning across turns, they generally fail to exhibit meaningful experiential learning across episodes, highlighting their inability to extract and utilize high level information from past experiences. 
\end{itemize}

\input{./largeboxes/example_dialogue}

%% file: largeboxes/example_dialogue.tex
\begin{figure}[ht]
    \centering
    \includegraphics[width=1.0\linewidth]{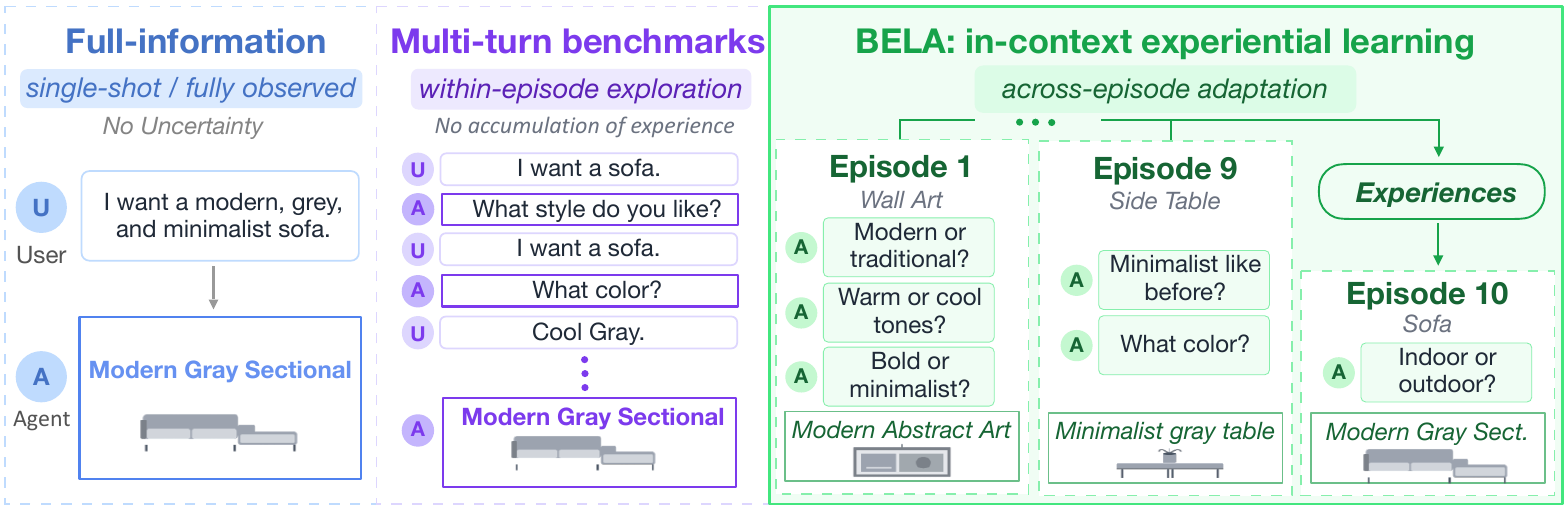}
    \caption{
    (\textbf{Left}) Typical agentic benchmarks (e.g. \citet{yao2024taubench}) focus on settings where all information is provided initially, and the model is tasked with producing the correct answer in a zero-shot fashion. 
    (\textbf{Middle}) Recent benchmarks (e.g. \citet{li2024mediq}) for LLM agents increasingly focus on multi-turn settings with uncertainty. 
    (\textbf{Right})
    Our In-Context Experiential Learning setting, focusing on across-episode adaptaion. 
    }
    \label{fig:setup}
\end{figure}

\definecolor{lightblue}{HTML}{3366CC}
\definecolor{cardblue}{HTML}{F7F9FF}
\definecolor{cardframe}{HTML}{4A4A4A}
\begin{figure*}[t]
\centering
\begin{tcolorbox}[
  enhanced,
  colback=blue!2!white,
  colframe=lightblue,
  arc=2mm,
  boxrule=0.8pt,
  title=\textbf{Recommendation Dialogues for In-Context Experiential Learning},
  coltitle=white,
  colbacktitle=lightblue,
  fonttitle=\bfseries\normalsize,
  halign title=center,
  left=1.5mm,
  right=1.5mm,
  top=1.5mm,
  bottom=1.5mm,
  boxsep=1mm
]
\footnotesize
\renewcommand{\arraystretch}{1.03}

\begin{minipage}[t]{0.318\linewidth}
\vspace{0pt}
\begin{tcolorbox}[
  enhanced,
  colback=cardblue,
  colframe=cardframe,
  arc=1.5mm,
  boxrule=0.5pt,
  left=1mm,
  right=1mm,
  top=1mm,
  bottom=1mm,
  equal height group=dialoguecards
]
\textbf{Ep 1: Hair Styling Gels}

\vspace{0.5mm}
\begin{tabularx}{\linewidth}{@{}>{\bfseries}p{13.5mm}@{\hspace{0.6mm}}X@{}}
AGT: & What is your hair type? \\
CUST 1: & Curly. \\
AGT: & What level of hold? \\
CUST 1: & Medium. \\
AGT: & What is your budget range? \\
CUST 1: & Around \$10 to \$15. \\
AGT: & Do you prefer a gel that provides moisture? \\
CUST 1: & I prefer a gel that [...] \\
\end{tabularx}

\vfill
\textbf{Selected Product}\\
ID: 34444\\
\emph{Frizz Free Styling Smoother}

\vfill
\textbf{CUST 1 Feedback}\\
I appreciate the suggestion, but I was actually looking for something more focused on care, like a conditioning balm or oil to keep it healthy and styled.
\end{tcolorbox}
\end{minipage}
\hfill
\begin{minipage}[t]{0.318\linewidth}
\vspace{0pt}
\begin{tcolorbox}[
  enhanced,
  colback=cardblue,
  colframe=cardframe,
  arc=1.5mm,
  boxrule=0.5pt,
  left=1mm,
  right=1mm,
  top=1mm,
  bottom=1mm,
  equal height group=dialoguecards
]
\textbf{Ep 2: Packaged Rice Dishes}

\vspace{0.5mm}
\begin{tabularx}{\linewidth}{@{}>{\bfseries}p{13.5mm}@{\hspace{0.6mm}}X@{}}
AGT: & What's your budget range for packaged rice dishes? \\
CUST 1: & I usually aim to keep it under \$10 per package. \\
AGT: & Do you have any dietary restrictions? \\
CUST 1: & I prefer a plant-based diet. \\
AGT: & Do you prefer specific rice varieties? \\
CUST 1: & I prefer wild rice. \\
\end{tabularx}

\vfill
\textbf{Selected Product}\\
ID: 72329\\
\emph{Dr.\ McDougall's Vegan Pilaf}

\vfill
\textbf{CUST 1 Feedback}\\
Oh, this looks great! I really appreciate the suggestion, but I usually prefer something with more whole grains.
\end{tcolorbox}
\end{minipage}
\hfill
\begin{minipage}[t]{0.318\linewidth}
\vspace{0pt}
\begin{tcolorbox}[
  enhanced,
  colback=cardblue,
  colframe=cardframe,
  arc=1.5mm,
  boxrule=0.5pt,
  left=1mm,
  right=1mm,
  top=1mm,
  bottom=1mm,
  equal height group=dialoguecards
]
\textbf{Ep 3: Almonds}

\vspace{0.5mm}
\begin{tabularx}{\linewidth}{@{}>{\bfseries}p{13.5mm}@{\hspace{0.6mm}}X@{}}
AGT: & Do you prefer raw or roasted almonds, or flavored varieties? \\
CUST 2: & Raw almonds. \\
AGT: & What are your preferences for this product category? \\
CUST 2: & I appreciate products that are sustainable, environmentally friendly, and ethically sourced. \\
\end{tabularx}

\vfill
\textbf{Selected Product}\\
ID: 72479\\
\emph{Roasted Sugared Almonds}

\vfill
\textbf{CUST 2 Feedback}\\
Oh dear, I was really hoping for something more in line with sustainable and locally sourced goods. This recommendation seems a bit off.
\end{tcolorbox}
\end{minipage}

\end{tcolorbox}

\caption{\textbf{Benchmark for Experiential Learning and Active Exploration (BELA).}
Example recommendation dialogues for in-context experiential learning across two customer personas and three choice sets.}
\label{fig:multi_episode_examples_unified}
\end{figure*}

%% file: sections/problem_setup.tex




Product recommendation provides a rich setting for testing the experiential learning capabilities of agentic systems.
Consider a common shopping scenario: a \textit{customer} ($c$) enters a store in seek of a product to satisfy a specific need.
Many products fulfill the same functional purpose, but they differ in style, aesthetics, or price, leading to preferences unique to the customer.
We refer to this group of functionally equivalent products as a \textit{choice set} ($\boldsymbol S$).
The recommender agent must discover customer's preferences over $\boldsymbol S$ by engaging in \textit{multiple turns} of queries: asking targeted questions and synthesizing the customer's responses.
This entire interaction, from initial inquiry to final recommendation, constitutes an \textit{episode}. 

Following each recommendation, the agent typically receives \textit{feedback}, such as a purchase decision or free-form text opinions about the suggested product.
This feedback, combined with the intermediate responses, encodes rich information about the \emph{latent factor} ($\boldsymbol \theta$) underlying a given episode. 
This latent factor may include the customer’s preferences, the dynamics of the question-answering process, how the customer perceives the structure of the choice set, and so forth. Compared to a typical sequential decision-making setup (e.g., POMDPs),
the ``reward" based on which the agent should optimize its strategy is encoded primarily in text as part of the observations \citep{yuksekgonul2025_textgrad}.
A capable agent must interpret these text-based signals to sharpen its belief of the latent $\boldsymbol{\theta}$ to improve its recommendations in subsequent episodes.

A performant recommender must learn across users and products and improve based on experience. Thus, we are not merely interested in the recommender agent’s performance within a single episode, but rather its ability to improve as its experience accrues.
For instance, consider a personalized recommendation setting where the same customer interacts with the agent across multiple episodes. In this case, the agent should focus on uncovering the customer’s underlying preferences to improve future recommendations.
In a scenario where the agent repeatedly sells a fixed choice set of products to a stream of new customers, the goal shifts to identifying how these products compare relative to one another across the diverse distribution of customers.
In the most difficult setting, both the customer and the choice set distributions can evolve over time. 

Formally, the agent encounters a \textit{stream of episodes} over its lifetime, where each $e$-th episode is characterized by a tuple $(c_e, \boldsymbol{S}_e)$ that induces a corresponding latent factor $\boldsymbol \theta_e$.
Each episode, indexed by $e \in  \{1, ..., E\}$, represents a single shopping session defined by a tuple $(c_e, \boldsymbol{S}_e)$, representing a customer and a choice set of products. 
In an episode (shopping session), customer $c_e$ is interested in buying one product from $\boldsymbol S_e$ according to their preference.
Each episode contains a sequence of turns where
each turn consists of the recommender's question or recommendation $a_{e,t}$, and the customer's response $o_{e, t}$. 
If the action $a_{e, t}$ is a question, 
then the response $o_{e, t}$ is the customer's answer;  if $a_{e, t}$ is a recommendation,
then $o_{e, t}$ is a feedback provided by the customer, encoding the customer's preference over the recommended product.
Naturally, the $e$-th episode concludes after a recommendation is made, and the agent moves on to the next episode. We denote this final feedback as $f_e$ for convenience.

We denote the  the sequence of latent factors by $\boldsymbol \Theta = \{\boldsymbol \theta_e |\ e=1,2,... \}$; some components of these latent factors may remain stable over time, while others may shift.
An intelligent agent must learn to adapt to the dynamics of $\boldsymbol \Theta$ in an online fashion.
At each turn, the response $o_{e, t} = g(\boldsymbol{\theta}_e, a_{e, t})$ is a function of both the action and the latent factor of the episode.
At episode $e$ and turn $t$, the entire history available to the recommender agent is \begin{align*}
    \mathcal H_{e, t} &= \underbrace{
        \bigl\{
            (a_{e', t'}, o_{e',t'}) 
            \vert\ 
            \forall e'\in [e-1], \forall t' \in [T_{e'}]
        \bigr\} 
    }_{\color{darkerblue} \displaystyle \text{Past Episodes}}
    \cup
    \underbrace{
        \bigl\{
        (a_{e, t'}, o_{e,t'}) 
        \vert\ 
        \forall t' \in [t-1]
        \bigr\} 
    }_{\color{darkerblue} \displaystyle \text{Current Episode}},
\end{align*}
with which the agent $\pi(\cdot)$ takes an action $a_{e, t} = \pi(\mathcal H_{e, t})$.

To measure the quality of the recommendations, we score the products for each unique pair of $(\boldsymbol{S}, c)$ that defines an episode.
Given one such pair, for each product $p_a \in \boldsymbol{S}$, we require a  score $y_a = h(p_a, c)$.
The main metric of merit for the recommendation is the \textit{regret}: $y^* - y_r$, where $y^*$ is a highest score, $y^* = \max\left( \{ y_a |\ y_a = h(p_a, c, \boldsymbol S )\ \forall p_a \in \boldsymbol S\} \right) $, and $y_r$ is the score of the recommended product.

%% file: sections/benchmark_setup.tex
\newcolumntype{C}{>{\centering\arraybackslash}X}






\begin{figure*}
    \centering
    \includegraphics[width=\textwidth, clip]{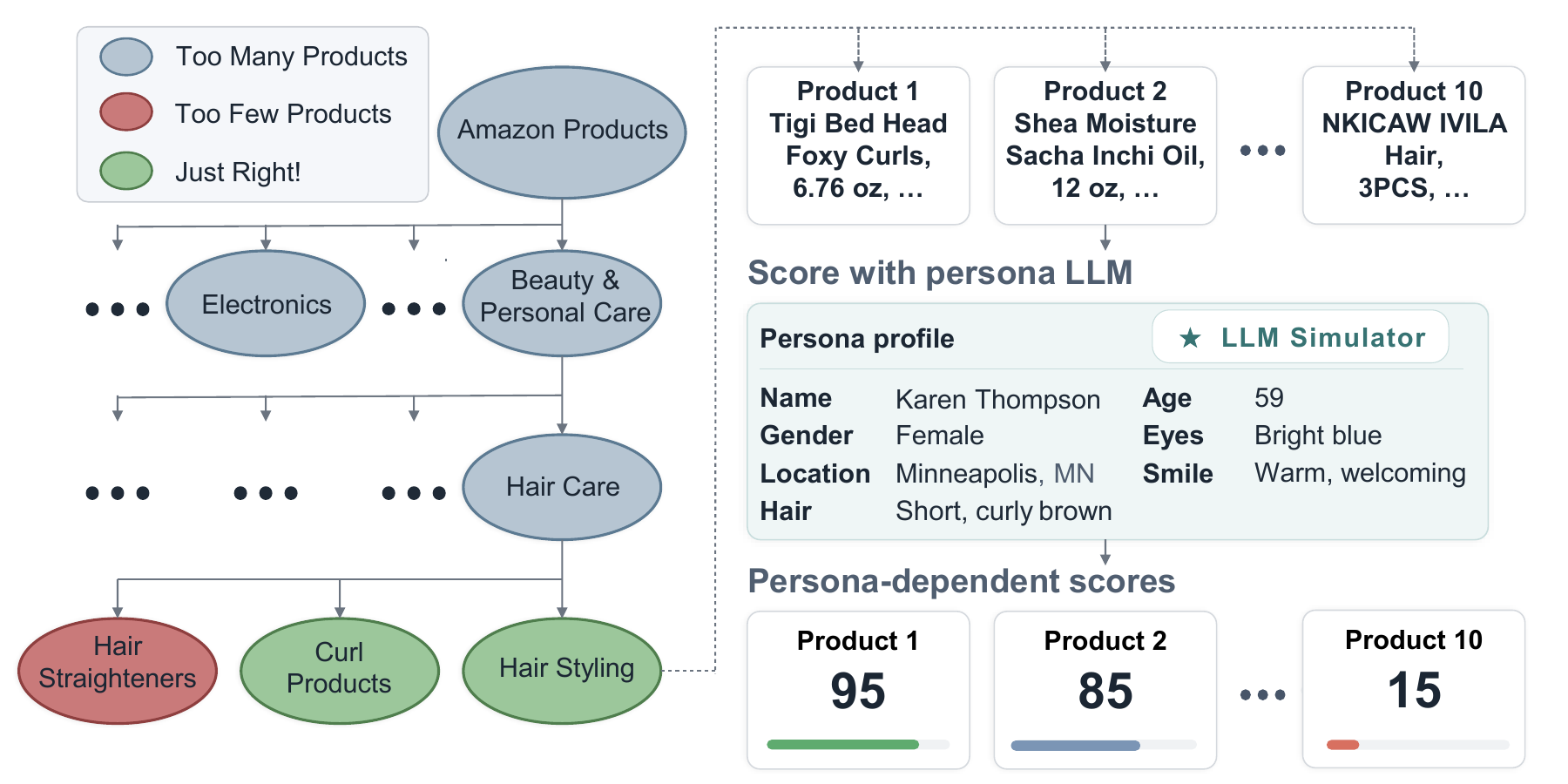}
    \caption{
        (\textbf{Left}) We use a predefined tree of categories \citep{spotlight_categories} and filtered the ones unsuitable as choice sets. We then score the products within each choice set with a persona-simulating LLM. An exemplar persona taken from \citet{li2025llmgeneratedpersonapromise} and their preferences over a category of products from \citet{hou2024_amazonproducts}. (\textit{Right}) Scoring is done by GPT-4o and Gemini-1.5-Pro. Consistent with the persona’s curly hair, curl-enhancing products are rated highly, whereas straightening products receive low scores.    
        }
    \label{fig:categories}
\end{figure*}

                  


To instantiate the aforementioned benchmark for experiential learning, our product recommendation dataset consists of two main modules: 
(1) a set of $\boldsymbol{C} = \{c_i\}$ of hypothetical customers that can respond \textit{interactively} to the recommender,
and 
(2) a set $\mathcal S = \{\boldsymbol{S}_j\}$ of choice sets.

A few desiderata are essential for a meaningful evaluation of experiential learning capability. 
First, the underlying sets of both modules ($\boldsymbol{C}$ and $\mathcal S$) need to be scalable and diverse to support a wide variety of $\boldsymbol{\Theta}$, customer preferences, and question-answering dynamics.
Second, the customer module must maintain \textit{consistency} between $\boldsymbol{\theta}_e$ and $(o_{e, t}, f_e)$; naturally, since the latter constitutes the experiences accessible to the agents, it must reliably inform the agents about $\boldsymbol{\theta}_e$. 
In particular, note that {whether $\boldsymbol{C}$ reflect the distribution of real-world customers' preferences \textbf{is not} a key requirement for the primary purpose of evaluating experiential learning}.
For use cases where one wishes to assess the quality of an agent's recommendation for customers from a particular industry or company, our framework easily supports customized instantiations of $\boldsymbol{C}$.

\subsection{Products and Choice Sets Curation}

To form the choice sets module, we curate a pool of products that are correctly categorized into reasonable choice sets.
For the products, we first sample 100K products from the Amazon Reviews dataset \citep{hou2024_amazonproducts}.
We then applied filters to remove products missing essential fields such as product descriptions. 
We match each product to a predefined category from \citet{spotlight_categories}. Through a rigorous matching process, we were able to assign 70\% of the sampled products to the categories of \citet{spotlight_categories};
see Appendix~\ref{app:cat_matching} for additional details.


Then, we extract $\mathcal S$ as a subset of predefined categories, leaving only categories that could reasonably be treated as choice sets. 
We first truncate the top two levels of this category hierarchy as these levels contained overly broad classifications ill-suited as choice sets  (e.g., \texttt{Home \& Kitchen}). 
We also filter categories that contain too few products to support nontrivial recommendation settings. 
See Figure~\ref{fig:categories} for an illustration of the process.

Through the novel filtering schemes, we construct a large collection of products and choice sets, consisting of 71K products and 2K choice sets.
Since some categories share products, the dataset contains in total 100K product-category pairs.


On scalability of available choice sets, 
our pipeline naturally supports expansion by sampling additional products from \citet{hou2024_amazonproducts} or other large-scale datasets.
Moreover, our framework is designed to be broadly compatible, allowing it to incorporate any choice set that contains a sufficient number of products with descriptions.
More sophisticated or specialized data collection pipelines could be developed to gather richer and more diverse choice sets, further extending the scope of our benchmark.


\subsection{Persona-Based Customer Simulation}
To power the customer module, we first need to curate a pool of persona descriptions that define the simulated customers.
Next, to enable the evaluation of recommendation quality in regret, we generate each persona’s preference over a given choice set ($y_a = h(p_a, c, \boldsymbol S)$).
Finally, we have a choice over the types of feedback $f_e$ to be provided at the end of a produce recommendation episode. 

\textbf{Persona Curation.}
To power the customer module, we use the 1M persona specifications in \citet{li2025llmgeneratedpersonapromise} to provide diversity in $\boldsymbol{C}$. 
We then utilize LLMs to simulate the customer's response corresponding to a persona file to enable interactive question-answering. 
The persona therein are generated by LLMs, and hence are as scalable and diverse as what modern LLMs encode.  

\textbf{Preference Generation.}
For preference generation ($y_a = h(p_a, c, \boldsymbol S)$), 
the scoring function $h(\cdot)$ was an average of the scores generated using two models: GPT-4o and Gemini-1.5-Pro.
Each model received the full description of the current persona encoding $c$, along with details of the products in the set $\boldsymbol{S}$ under evaluation. 
The models were instructed to assign a score (0–100) reflecting how much the persona would enjoy or appreciate each product within $\boldsymbol{S}$. 
The consistency of the scoring process is crucial to the integrity of the benchmark.
To evaluate the consistency, we randomly selected five categories and measure the average variation in the scores across three seeds. 
We found that the averaged score variation is $4.1$, much smaller than the standard deviation in scores within each category ($25.4$), which suggests that the scoring process is consistent.
For instance, in Figure \ref{fig:categories}, the persona is described as having \emph{curly brown hair}. Consistent with this trait, products intended for curl enhancement are assigned high scores, while products designed for hair straightening receive lower scores.

\textbf{Customer Simulation and Feedback.}
Users are simulated by a LLM (GPT-4o) prompted to respond to agent's questions as the persona when shopping within a given product category. 
At the end of each $e$-th episode, the simulated user provides a feedback $f_e$. 
We support three forms of feedback: (1) \textbf{Regret}, (2) \textbf{Stars}, and (3) \textbf{Free-form Text}. 
Regret feedback directly provides the regret of the recommendation. While informative, this metric is unrealistic in practice, as it is typically infeasible to obtain exact utility scores from real users. 
To approximate more practical feedback, we provide star ratings as $5\times\frac{\text{score of chosen product}}{\text{score of best product}}$, rounded to the nearest integer, mirroring the star systems commonly found on e-commerce platforms. 
Finally, free-form text feedback involves prompting the customer LLM with the recommendation, its regret value, and the top three scored products, and instructing it to generate natural language feedback—expressed in the persona’s voice on the extent to which the chosen product meets their preferences. 
See Figure~\ref{fig:multi_episode_examples_unified} for examples.
The generated feedback reflects the persona’s curly hair type and consistently maintains the persona’s character throughout the interaction.
We provide additional experiments for the consistency of module in Appendix~\ref{app:consistency}.

\subsection{Dataset Robustness}
To prevent unrealistic scenarios (e.g., an individual with no musical interests shopping for pianos), we exclude choice set-persona pair in which the highest product score is below 60. 
Using this filtering procedure, an average of $3.75$ scenarios were skipped across 10 randomly drawn choice set-persona pairs. 

We further verify that the customer module forms a consistent and learnable environment through two additional checks: (a) faithfulness of simulator responses to persona descriptions, and (b) stability of simulator responses under repeated queries. For (a), we sample 10 multi-episode trajectories and use Claude Sonnet~4 as an LLM-as-judge to label each persona-agent answer as \emph{Aligned}, \emph{Conflict}, or \emph{Neutral} relative to the persona description. Across 327 answers, 74.6\% are labeled \emph{Aligned}, 19.3\% \emph{Neutral}, and 6.1\% \emph{Conflict}. Manual review shows that all labeled conflicts are false positives, arising from missing persona details, unwarranted assumptions by the judge, or differences in specificity  than genuine contradictions. For (b), we resample the simulator's answer ten times for each question and measure agreement with the per-question majority response. Across 1{,}090 responses, 94.4\% agree with the majority position, suggesting high response stability. Full setups and results are in Appendix~\ref{app:consistency}.

To further demonstrate learnability of the environments, we conducted a manual, human-curated questioning run as seen in Appendix \ref{app:manual}. The questions were deliberately \emph{reasonable} in that they (i) targeted concrete, product-determining attributes, (ii) prompted the persona to reveal personality details, and (iii) were answerable without specialized knowledge. 
Using the information elicited by expert questionings, the agent identified the best scoring product (regret = 0) when it had otherwise consistently chosen a product with regret 37.5. This demonstrates that, given a well-planned query policy, the information required to reach the best item is accessible, and the environment is \emph{solvable}. 


%% file: sections/experiments.tex
We consider three sequential recommendation settings that exemplify experiential learning:
(a) same customer across episodes with different choice sets each episode, 
(b) same category across episodes with different customers each episode, 
and (c) different choice sets and customers each episode. 
Of course, the rich sets of personas and choice set available in BELA can support many more settings than the ones considered in this study. 

To evaluate performance, aside from the aforementioned regret metrics, we monitor the number of questions asked.
In our experiments, recommender agents are instructed to identify the best product while asking as few questions as they deem necessary. This setup requires agents to balance uncertainty reduction with interaction efficiency: they should ask enough questions to make an informed recommendation, but avoid unnecessary questions. We consider three baselines throughout.
\textsc{Random}  recommends products uniformly at random, and \textsc{Popularity} selects the product with the highest original rating, breaking ties by choosing the cheaper option. 
Finally, \textsc{Oracle} is a Claude-Sonnet-4 with access to the full persona description and prompted to directly select a product from the category. Since Claude is not used during the initial scoring of products, this setup ensures a clean separation and serves as a lower bound on regret in cases where the recommender has full access to persona information.

\subsection{Do Models Learn From Experiences?}
\label{sec:exp-model-fail}
We evaluate a broad suite of frontier and open-weight models, including GPT-5.4, Gemini-3.1-Pro-Preview, Claude-Opus-4.6, Kimi-K2, Qwen-Max, and DeepSeek-Reasoning. In the main text, we report representative results for three frontier models: GPT-5.4, Gemini-3.1-Pro-Preview, and Claude-Opus-4.6 over $10$ episodes across $40$ random seeds under setting (a), using Free-form Text feedback. Results for the remaining models are provided in Appendix~\ref{app:ablation}.

\begin{figure}[ht]
    \centering
    \includegraphics[width=\linewidth]{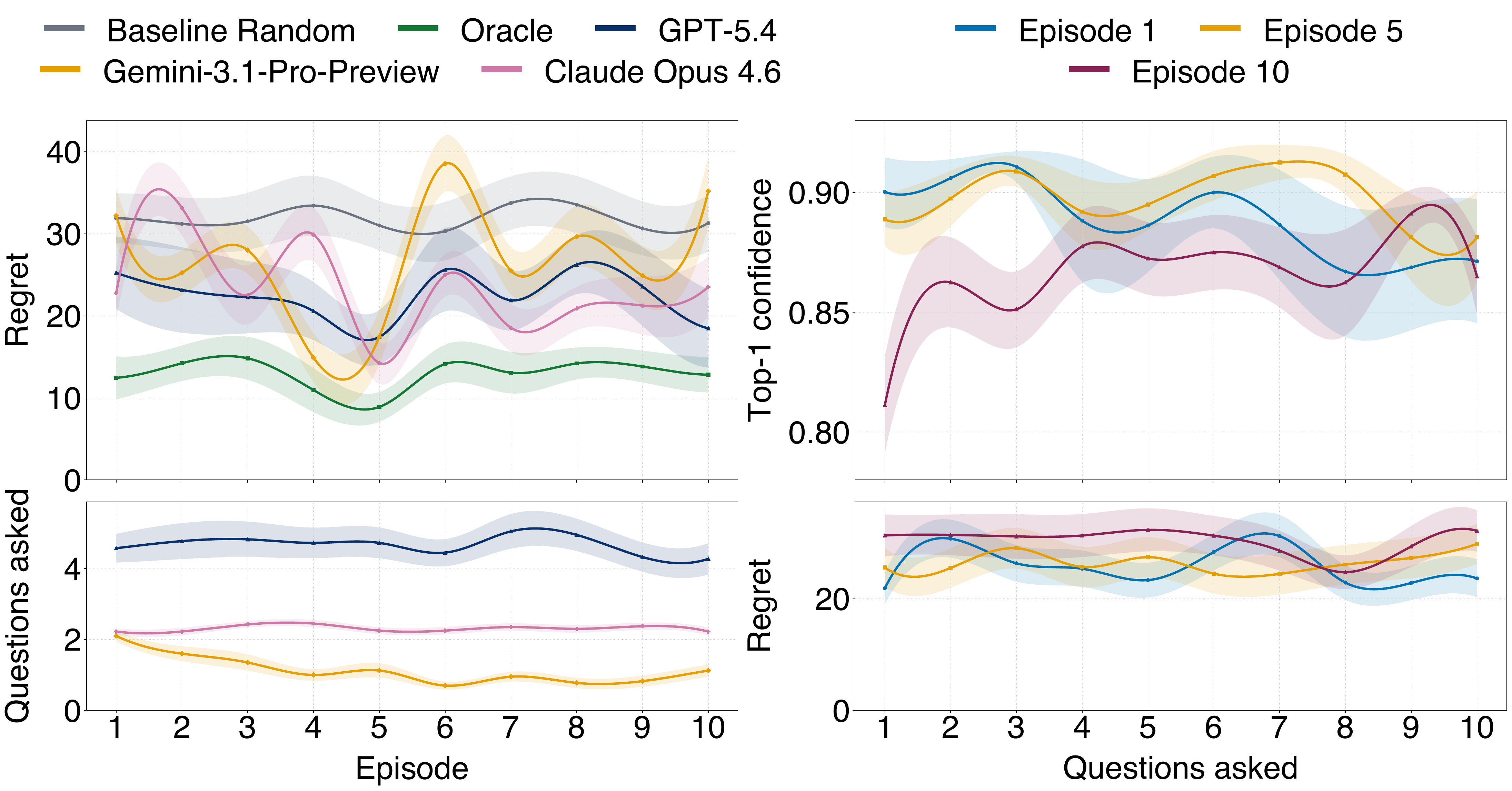}
    \caption{(\textbf{Left}) All models perform worse than the oracle and do not demonstrate learning across episodes. Models ask either the same or fewer questions with more episodes. 
    (\textbf{Right}) GPT-5.4 maintains a similar level of top-1 confidence across episodes; however, shows no improvement as more questions are asked. 
    }
    \label{fig:results1}
\end{figure}

As shown in Figure~\ref{fig:results1}, while the models perform better than simple baselines, all significantly underperform the oracle baseline. 
In particular, no models exhibit episode-over-episode improvements, highlighting the difficulty for current SOTA models to leverage experiences about the customer obtained in prior episodes. 

If agents perform poorly, we expect them to ask more questions to improve their recommendations. 
However, 
we find that the number of questions asked generally remains the same for most models, and even declines over episodes for Gemini-2.5-Pro and Gemini-2.5-Flash.
We highlight this lack of proactive information gathering as a major deficiency in the behavior patterns of SOTA models. 

Overall, across various recommendation settings, model families, agentic scaffolds, feedback types, prompts etc., the models fail to demonstrate experiential learning capability.
We provide the results of these ablation studies in Appendix \ref{app:ablation}.

\definecolor{lightblue}{HTML}{3366CC}
\definecolor{cardblue}{HTML}{F7F9FF}
\definecolor{cardframe}{HTML}{4A4A4A}

\begin{figure*}[t]
\centering
\begin{tcolorbox}[
  enhanced,
  colback=blue!2!white,
  colframe=lightblue,
  arc=2mm,
  boxrule=0.8pt,
  title=\textbf{Examples from the Ideal Trajectory of Experiential Learning},
  coltitle=white,
  colbacktitle=lightblue,
  fonttitle=\bfseries\normalsize,
  halign title=center,
  left=1.5mm,
  right=1.5mm,
  top=1.5mm,
  bottom=1.5mm,
  boxsep=1mm
]
\footnotesize
\renewcommand{\arraystretch}{1.03}

\begin{minipage}[t]{0.34\linewidth}
\vspace{0pt}
\begin{tcolorbox}[
  enhanced,
  colback=cardblue,
  colframe=cardframe,
  arc=1.5mm,
  boxrule=0.5pt,
  left=1mm,
  right=1mm,
  top=1mm,
  bottom=1mm,
  equal height group=idealtrajcards
]
\textbf{Ep 1}

\vspace{0.5mm}
\begin{tabularx}{\linewidth}{@{}>{\bfseries}p{9mm}@{\hspace{0.3mm}}X@{}}
AGT:  & Do you prefer solar-powered or low-voltage lights? \\
CUST: & I prefer solar-powered lights because they align with my sustainability values. \\
AGT:  & Do you prefer warm white or color-changing lights? \\
CUST: & Warm white---they create a cozy and inviting atmosphere. \\
\end{tabularx}
\end{tcolorbox}
\end{minipage}
\hfill
\begin{minipage}[t]{0.025\linewidth}
\vspace{0.75in}
\makebox[0pt][c]{\hspace{4mm}\Large$\cdots$}
\end{minipage}
\hfill
\begin{minipage}[t]{0.27\linewidth}
\vspace{0pt}
\begin{tcolorbox}[
  enhanced,
  colback=cardblue,
  colframe=cardframe,
  arc=1.5mm,
  boxrule=0.5pt,
  left=1mm,
  right=1mm,
  top=1mm,
  bottom=1mm,
  equal height group=idealtrajcards
]
\textbf{Ep 6}

\vspace{0.5mm}
\begin{tabularx}{\linewidth}{@{}>{\bfseries}p{9mm}@{\hspace{0.3mm}}X@{}}
AGT:  & Would you consider a product outside your price range if it meets your sustainability and quality criteria? \\
CUST: & Yes, I would---quality and sustainability matter more to me than price. \\
\end{tabularx}
\end{tcolorbox}
\end{minipage}
\hfill
\begin{minipage}[t]{0.025\linewidth}
\vspace{0.75in}
\makebox[0pt][c]{\hspace{4mm}\Large$\cdots$}
\end{minipage}
\hfill
\begin{minipage}[t]{0.25\linewidth}
\vspace{0pt}
\begin{tcolorbox}[
  enhanced,
  colback=cardblue,
  colframe=cardframe,
  arc=1.5mm,
  boxrule=0.5pt,
  left=1mm,
  right=1mm,
  top=1mm,
  bottom=1mm,
  equal height group=idealtrajcards
]
\textbf{Ep 10}

\vspace{0.5mm}
\begin{tabularx}{\linewidth}{@{}>{\bfseries}p{9mm}@{\hspace{0.3mm}}X@{}}
AGT:  & Do you prefer mules and clogs made from natural materials like leather or canvas? \\
CUST: & Yes---natural and sustainable materials fit my values best. \\
\end{tabularx}
\end{tcolorbox}
\end{minipage}

\end{tcolorbox}

\caption{\textbf{Ideal Trajectory of Experiential Learning.}
Illustration of how an agent should evolve across episodes in the personalization setting. Here the agent progressively identifies the customer's preference for natural and sustainable materials.}
\label{fig:ideal_traj}
\end{figure*}

\subsection{Can Models Comprehend Their Uncertainties?}

\begin{figure}[t]
\centering

\begin{minipage}[c]{0.67\linewidth}
    \centering
    \includegraphics[width=\linewidth]{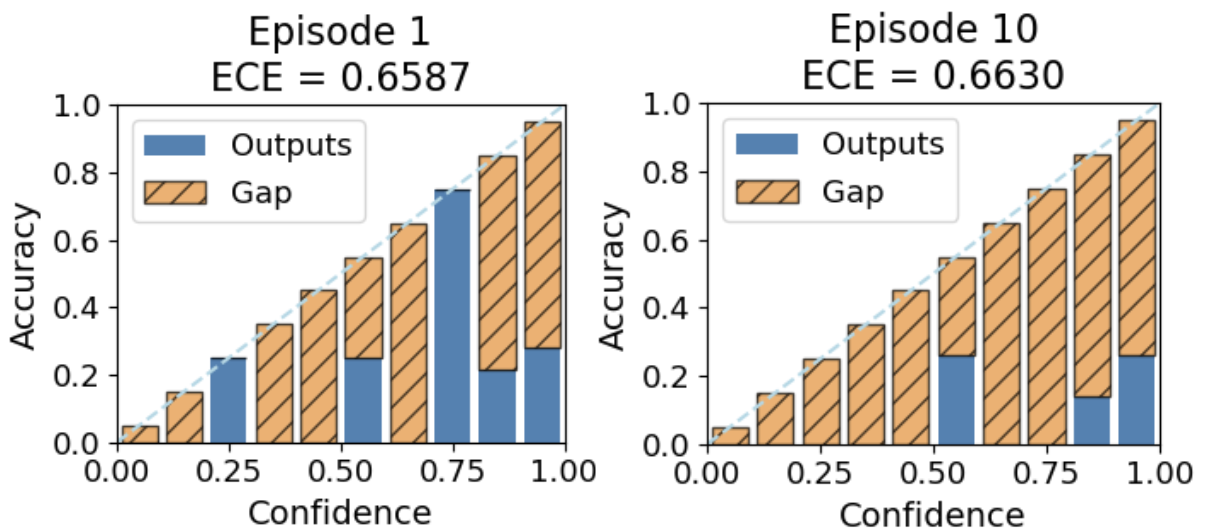}
\end{minipage}
\hfill
\begin{minipage}[c]{0.31\linewidth}
    \footnotesize
    \captionof{figure}{\textbf{ECE plots for GPT-4o.}
    GPT-4o remains poorly calibrated from Episode~1 to Episode~10.
    Confidence is measured on whether regret is within 10 points across turns, suggesting that in-context experience does not reliably improve uncertainty estimates.}
    \label{fig:ece_abridged}
\end{minipage}

\end{figure}
If already defined earlier (e.g., in Fig. 2), you do not need to redefine these.

Next, we examine whether agents can better quantify their own uncertainties from accrued experiences. 
To assess this, in the otherwise same setting as in Section~\ref{sec:exp-model-fail}, we prompt GPT-4o to output six kinds of confidences on: (1-2) whether the chosen item would fall into top-1 and top-5 favorite products of the customer, and (3-6) whether the realized regret would fall into the expected regret within 5, 10, 20, and 30 points. 
In all experiments, the agent was run for 10 episodes under setting (a), with additional evaluations conducted in the 1st, 5th, and 10th episodes:
in these episodes, the agent was additionally prompted to make a recommendation after each question, allowing us to measure regret and confidence at every turn. 

Figure~\ref{fig:results1} shows that the GPT-4o seems to grow more confident in episode 10. 
However, the agent also does not show any improvement in regret across episodes. 
We also find that the model struggles to estimate its uncertainties for all episodes (see Figure~\ref{fig:ece_abridged}). 
We provide complete experimental results for all confidences considered in Appendix~\ref{app:ece}, where the model consistently fails to calibrate from experiences.

\subsection{Ideal Learning Trajectories}
\begin{figure}[t]
    \centering
    \includegraphics[width=\linewidth]{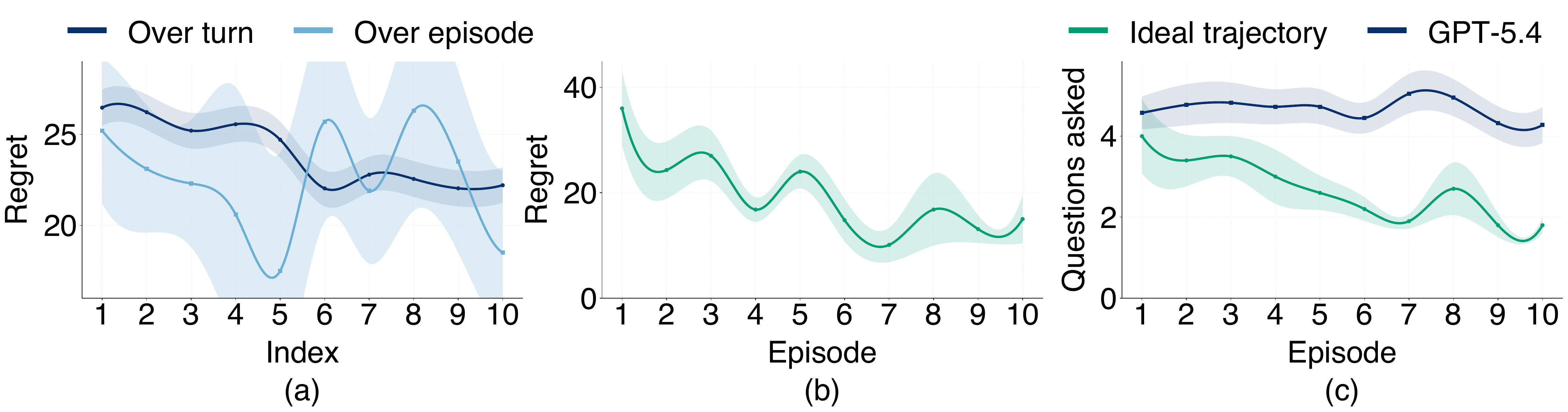}
    \caption{
        \textbf{(a)} GPT-5.4 exhibits within-dialogue learning across turns, but does not improve across episodes. \textbf{(b)} In contrast, ideal trajectories obtained from sampling good questions demonstrated possibility to improve across episodes by gathering and reusing useful information from prior interactions. 
        \textbf{(c)} The ideal trajectories ask progressively fewer questions over episodes, as one would expect for proficient recommender learning about their customers. 
        In contrast, GPT-5.4 asks a roughly constant number of questions. Error bars denote standard errors.
    }
    \label{fig:turn_vs_episode}
\end{figure}

The negative results above admit a trivial alternative explanation: the customer interactions may carry little transferable information for models to learn across episodes. We rule this out by constructing ideal learning trajectories that exploit cross-episode information in the same experimental setting as Figure~\ref{fig:results1}. Their existence serves as a witness to shared information that agents could have leveraged but did not.
To find the ideal trajectories, for each $k$-th episode, we run the experiment 5 times with GPT-5.4 as the recommender. 
Out of the 5 repetitions, we pick one high-quality trajectory that contains questions that target meaningful aspects of the persona. 
We then freeze this accepted episode and use the questions and information obtained therein as the starting context for the next episode. 
We repeat this process for all 10 episodes to find an ideal trajectory, and in total find 10 trajectories depicted in Figure~\ref{fig:turn_vs_episode}. 
The decreasing regret across episodes illustrates how a capable agent \emph{should} inquire transferable questions to acquire experiences that help recommend products in later episodes. 
In contrast, without this deliberate process, current agents on their own appears incapable of this learning.

Qualitatively, these trajectories show a shift in focus from surface-level, episode-specific questions that focus on specific features, to higher-level modeling of user preferences that may be transferable. 
Figure~\ref{fig:ideal_traj} gives one representative example from Episodes~1, 6, and 10. In Episode~1, the agent asks shallow feature-comparison questions such as “solar-powered or low voltage?”, which captures surface-level preferences helpful to the recommendation in the current trajectories, but miss deeper latent preferences of the user. 
By Episode~6, the agent begins to model higher-level attributes such as the user’s value hierarchy (e.g., sustainability vs.\ price), signaling a transition from inquiring specific attributes to modeling implicit preferences. 
By Episode~10, the agent has internalized the persona’s consistent emphasis on natural materials and sustainability, and begins posing \emph{confirmation-style} questions that reflect an internal latent model of what the persona cares about.

%% file: sections/related_works.tex
\textbf{Reinforcement learning for LLM.} 
Reinforcement Learning with Human Feedback (RLHF) \citep{christiano_deep-rlhf, stiennon_rlhf, ziegler2020finetuninglanguagemodelshuman} has proven highly effective for aligning models with objectives that lack well-defined, differentiable reward functions, improve models on dimensions such as translation quality \citep{ramos_rlhf-translation, kreutzer_rlhf-translation}, helpfulness \citep{ouyang_rlhf, openai2024gpt4technicalreport}, and factual accuracy \citep{bai2022constitutionalaiharmlessnessai, glaese2022_rlhf-safety, touvron2023llama2openfoundation, sun_rlhf-factual}. 
However, these settings are fundamentally limited as RL environments, corresponding to fully observable single-step environments, where the model takes one action (i.e. generating a response) with all necessary information available, and immediately receives a scalar reward.
While this paradigm has been remarkably successful for training performant zero-shot models, it fails to capture the interactive, dynamic nature of real-world environments, where actions and feedback over multiple turns and episodes can aid the models' decisions.
In contrast, our work focuses on evaluating agents ability to contend with uncertainty in \textit{multi-turn} and \textit{multi-episode} settings, and highlights the insufficiency of current state-of-the-art models to adapt over the latter.

\textbf{Multi-turn and Multi-episode RL.}
Recent years have witnessed a surge of interest in agentic models powered by LLMs \citep{jimenez2024swebench, yao2024taubench, karten2025pokeagent}.
These agents are expected to autonomously plan, act and adapt through iterative interactions with their environment, requiring models to engage in \textit{multi-turn} interactions within an episode of task.
As a result, prior works have explored multi-turn settings in text-based games \citep{abdulhai2025_lmrl,tajwar2025paprika}, medical question-answering \citep{li2024mediq}, and numerous other tasks \citep{wang-scienceworld, liu2024agentbench}. 
Some other works have explored multi-episode settings \citep{wu2024streambench, zheng2025lifelongagentbenchevaluatingllmagents}, but they primarily focus on environments devoid of uncertainties, and therefore do not require agents to actively resolve uncertainties in the environment. 
In contrast, our setup focuses on assessing the ability of the models to \textit{reason through uncertainties} of the latent preference of the customers in context. 





\textbf{Recommendation System and LLM Personalization.} 
There has been a long line of work on recommendation systems \citep{resnick_cf, koren_mf}.
In the era of foundation models, LLMs have demonstrated remarkable zero-shot performance in recommendation tasks \citep{gengP5_recsys, He_llm-rec, lyu_benchmark-llm-rec}.
However, the classical yet crucial setting of sequential recommendation \citep{hidasi_session-rec, tan_session-rec} remains underexplored in this era, and the even more realistic problem of modeling interactions with customers has received little attention.
Our dataset addresses this gap by enabling the study of these settings through an interactive user simulator powered by LLMs.
Separate but relatedly, the rise of LLMs has sparked growing interest in personalized LLMs \citep{castricato_persona, jang2024personalized, zollo2025personalllm}.
Compared to these post-hoc heuristic approaches, we propose a more principled way to instill agents with the ability to learn to personalize.
Our dataset is designed to directly evaluate and help advance this capability.





%% file: sections/conclusion.tex
We argue for a shift in focus toward measuring whether agents can leverage past episodes to refine how they resolve uncertainty.
To this end, we introduced \textbf{BELA}: a flexible interactive recommendation benchmark designed to evaluate in-context experiential learning capabilities.
The framework we employ provides a uniquely scalable dataset, supporting up to 2B multi-turn product recommendation environments, the combinations of which can form numerous diverse environments to study experiential learning behaviours. 
Our experiments reveal that while frontier models can learn across turns within an episode, they still fail to learn effectively from experiences accrued across episodes.
In addition, they struggle to calibrate their uncertainties and, as a result, fail to proactively ask informative questions. 
By uncovering these limitations, we hope to motivate future research on experiential learning, which is essential for building agents capable of reliable decision-making under constantly shifting real-world uncertainties.
In terms of limitation, BELA deliberately prioritizes evaluation of experiential learning over realism, and we do not expect the simulated customers to reflect real users. 
This work also does not explore training methods to instill experiential learning capability, which we leave as interesting direction for future work.

%% file: sections/appendix.tex
\section{Ablation Studies}
\label{app:ablation}
In this section, we discuss additional ablation studies performed using the dataset.
Overall, SOTA models do not exhibit the ability to learn across episodes.

\subsection{Varying User Feedback}

\begin{figure}[H]
\centering

\begin{minipage}[c]{0.70\linewidth}
    \centering
    \includegraphics[width=\linewidth]{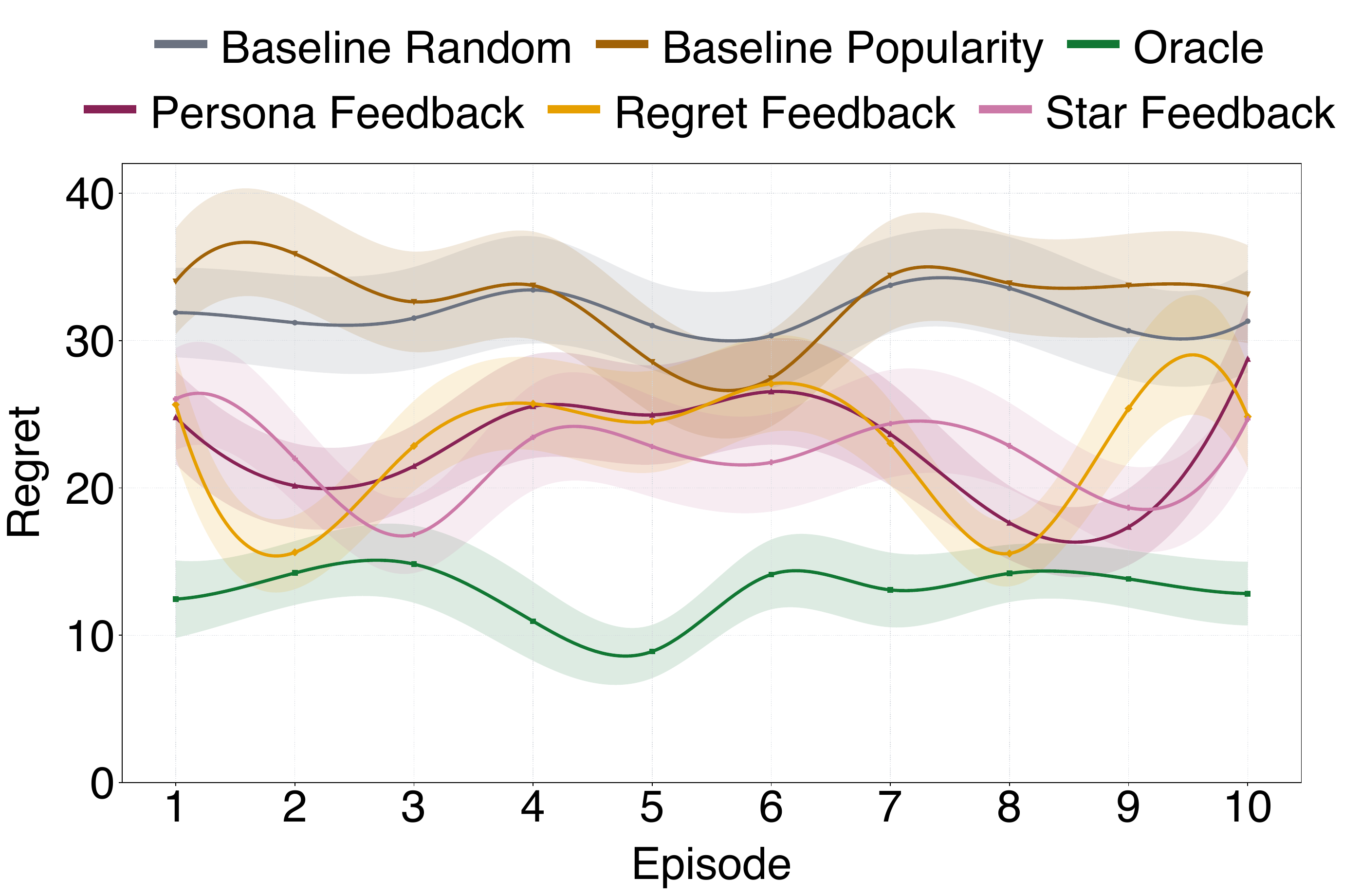}
\end{minipage}
\hfill
\begin{minipage}[c]{0.29\linewidth}
    \footnotesize
    \captionof{figure}{\textbf{Effect of feedback type.}
    Feedback type does not significantly affect model performance across episodes, suggesting that GPT-4o does not reliably use more informative customer feedback.}
    \label{fig:feedback}
\end{minipage}

\end{figure}

Fixing the model to GPT-4o and holding all else constant, we evaluated the three feedback types described in Section 4.3 across 40 random seeds and 10 episodes. As shown in Figure~\ref{fig:feedback}, no statistically significant differences were observed across feedback types.
Note that the Regret feedback type represents a less noisy source of information for the models.
The fact that no models seem to utilize the more informative feedback indicates the models are not able to learn from customer feedback.

\subsection{Varying Prompts}
We investigate whether explicit prompting can induce more deliberate planning during recommendation.
Using GPT-4o, we compare three prompting variants:
(1) \textbf{No Strategy}, which simply instructs the agent to ask questions for $T$ turns;
(2) \textbf{Greedy}, which instructs the agent to choose actions that maximize immediate recommendation quality; and
(3) \textbf{DP}, which frames the task as planning in a POMDP and instructs the agent to act as a DP-optimal planner.
The full prompts are provided in Appendix~\ref{app:planning_prompts}.

\begin{figure}[H]
    \centering
    \includegraphics[width=\linewidth]{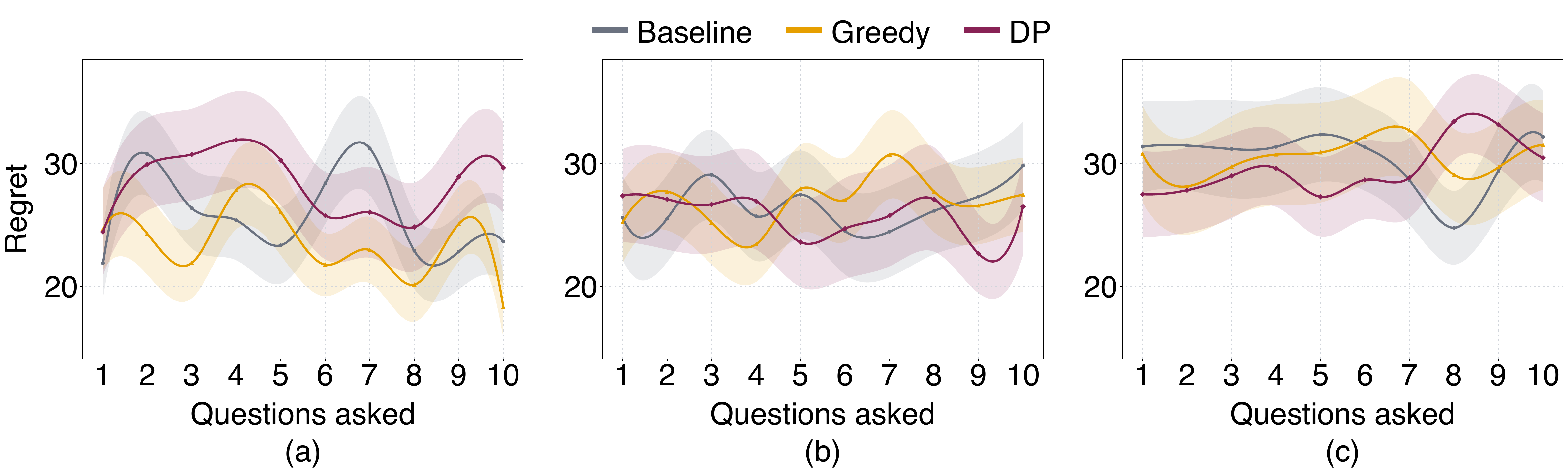}
    \caption{
    \textbf{Prompting strategies across turns.}
    Across 60 seeds, explicit greedy and DP-style planning prompts do not reliably improve GPT-4o's performance over turns in Episodes 1, 5, and 10.
    }
    \label{fig:planning}
\end{figure}

We first evaluate whether these prompting strategies improve within-episode behavior across turns.
In this setting, the agent is forced to ask 10 questions and to provide a recommendation after each question.
However, the agent receives no feedback between turns and is not given memory of its previous recommendations until the final turn.
We evaluate this setup in Episodes 1, 5, and 10.
As shown in Figure~\ref{fig:planning}, explicit planning prompts do not reliably improve the agent's ability to learn across turns.

\begin{figure}[H]
\centering

\begin{minipage}[c]{0.70\linewidth}
    \centering
    \includegraphics[width=\linewidth]{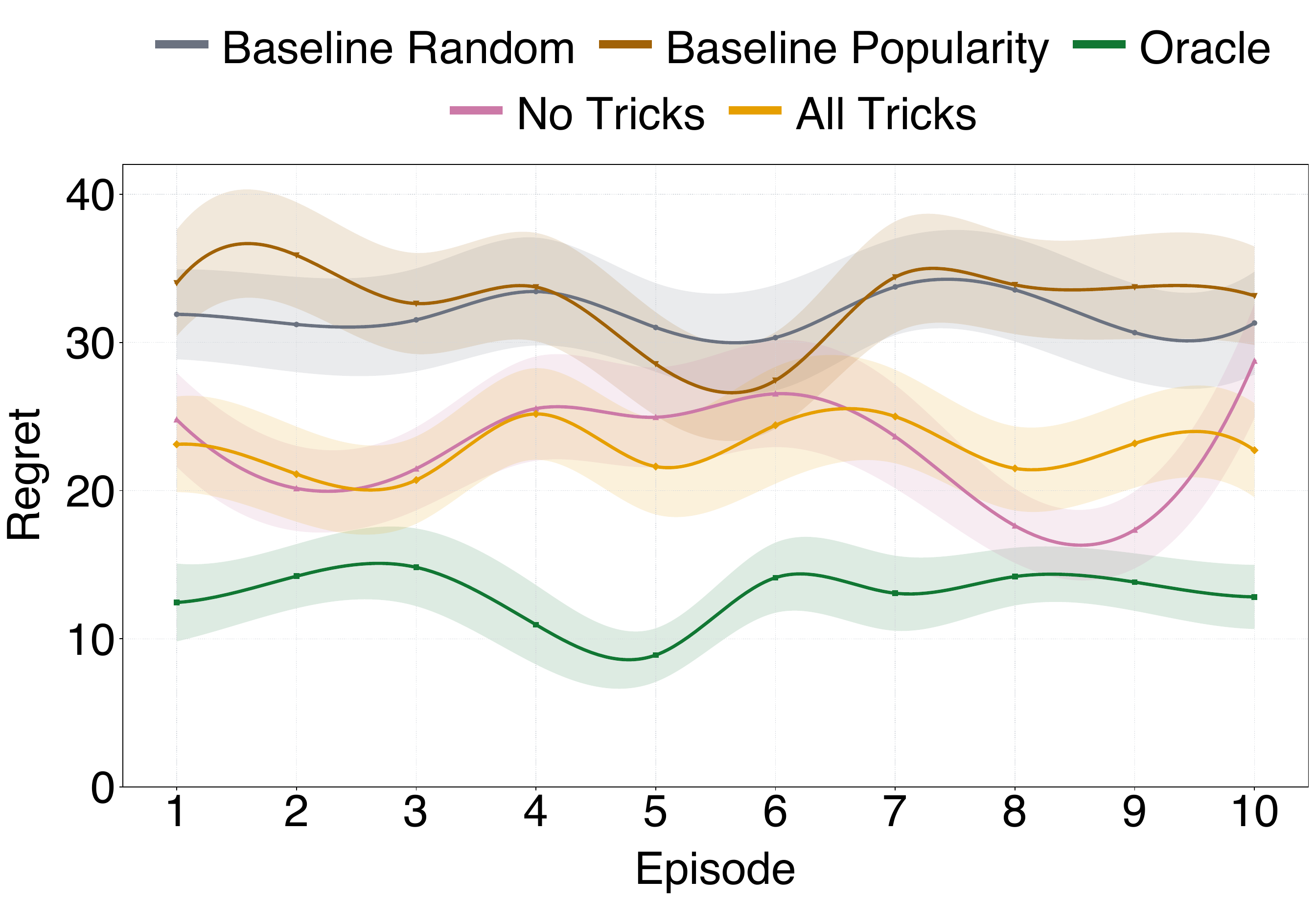}
\end{minipage}
\hfill
\begin{minipage}[c]{0.29\linewidth}
    \footnotesize
    \captionof{figure}{\textbf{DP prompting across episodes.}
    DP-style prompting does not significantly improve performance over the no-strategy baseline across the full episode sequence.}
    \label{fig:tricks}
\end{minipage}

\end{figure}

Furthermore, when investigating the DP-style prompt across the full 10-episode sequence, we find there is no statistically significant improvement over the no-strategy baseline as shown in Figure~\ref{fig:tricks}.
This suggests that simply describing an optimal planning strategy in the prompt is insufficient for inducing robust experiential learning.

\subsection{How to Utilize Past Experiences}
\begin{figure}[H]
\centering

\begin{minipage}[c]{0.70\linewidth}
    \centering
    \includegraphics[width=\linewidth]{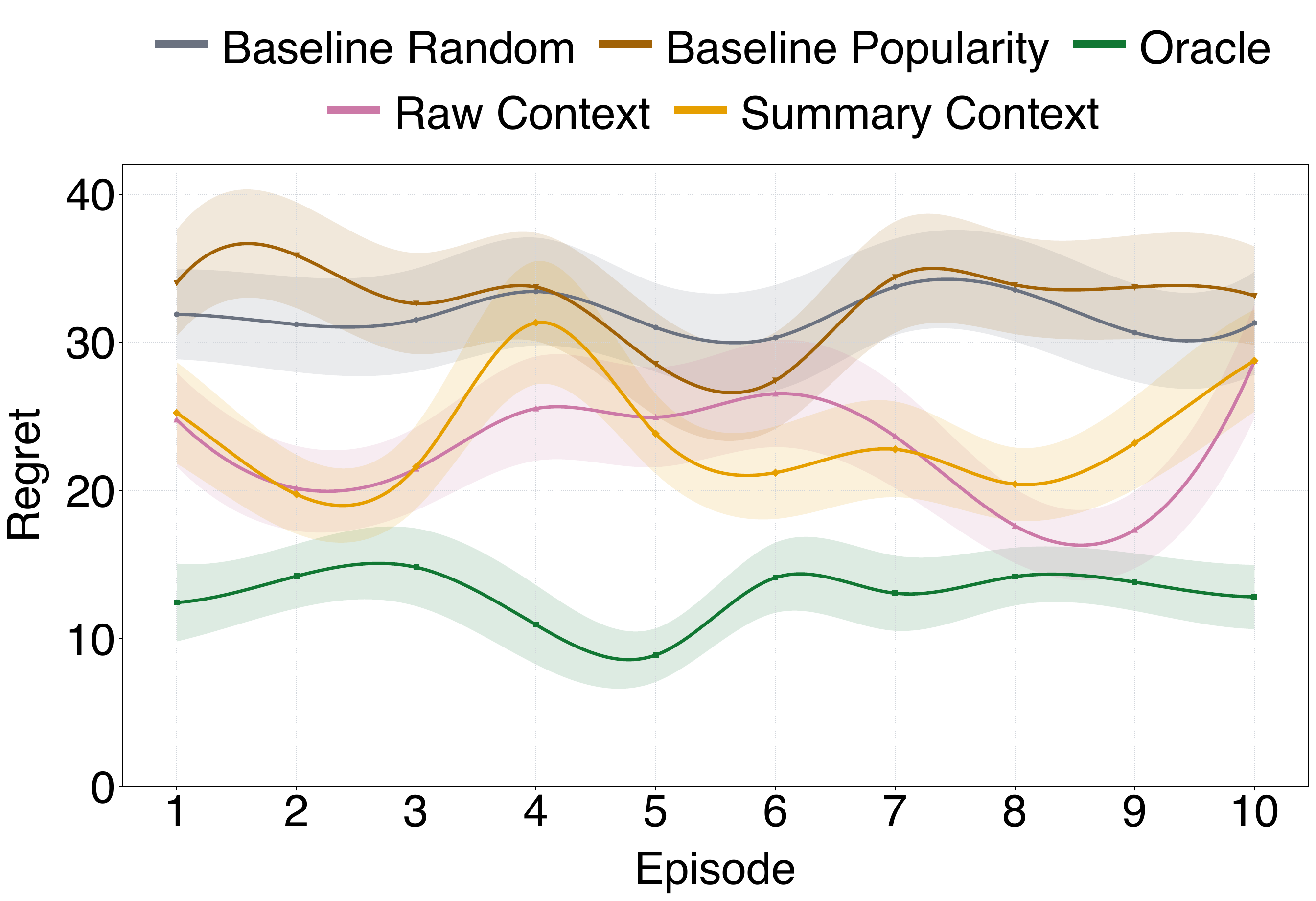}
\end{minipage}
\hfill
\begin{minipage}[c]{0.29\linewidth}
    \footnotesize
    \captionof{figure}{\textbf{Effect of context summaries.}
    Carrying forward agent-generated summaries of past episodes does not significantly improve GPT-4o's performance across episodes.}
    \label{fig:context}
\end{minipage}

\end{figure}

Fixing the model to GPT-4o and holding all else constant, we examine whether giving agents access to summaries of their past interactions improves performance.
In this variant, after each episode, the recommender agent generates a context summary that is carried forward into subsequent episodes.
As shown in Figure~\ref{fig:context}, this summary-based memory does not produce a statistically significant improvement, suggesting that simply preserving past interaction summaries is insufficient for robust experiential learning.

\subsection{Learning Across Different Users}
\begin{figure}[H]
\centering

\begin{minipage}[c]{0.70\linewidth}
    \centering
    \includegraphics[width=\linewidth]{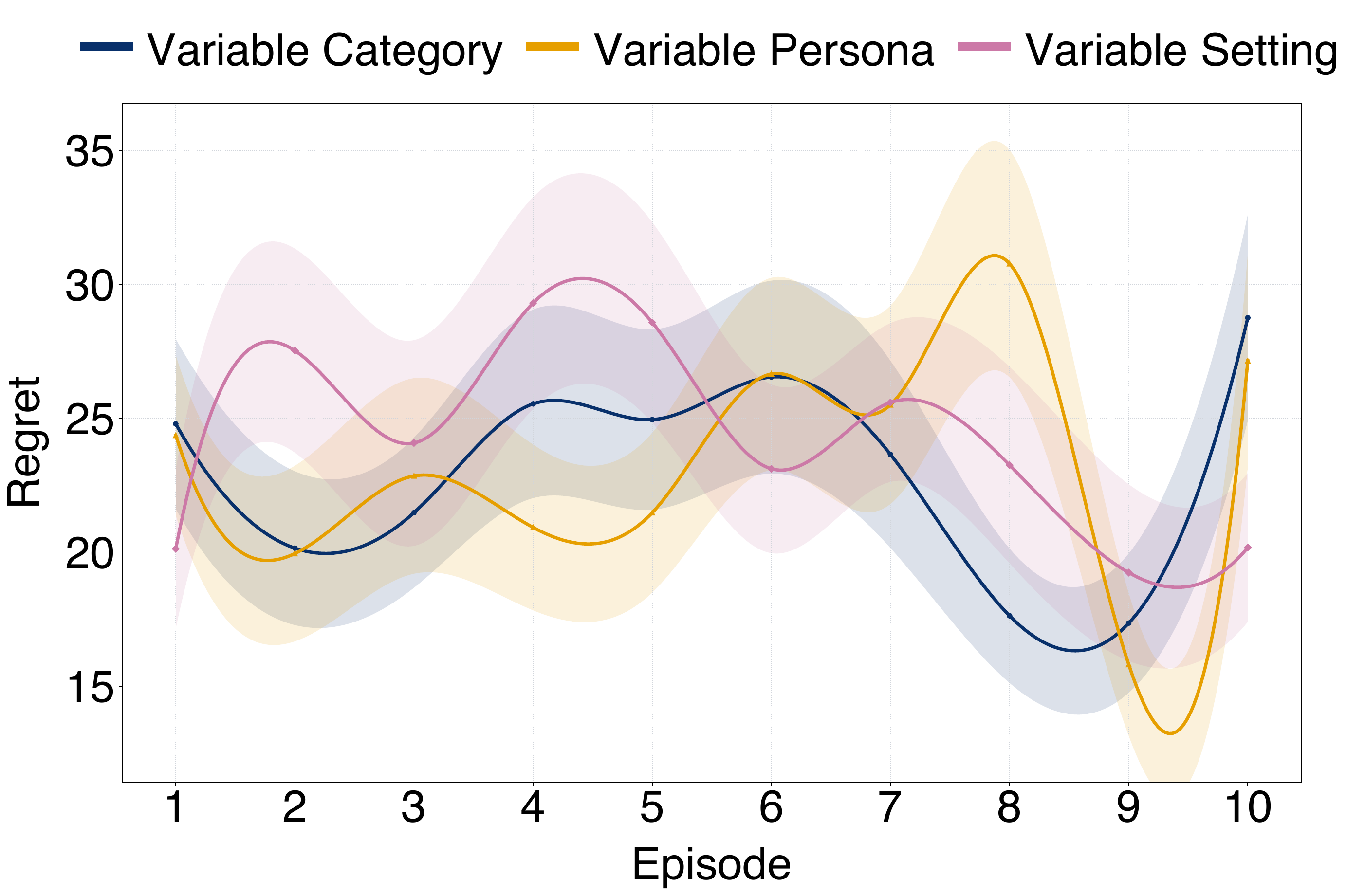}
\end{minipage}
\hfill
\begin{minipage}[c]{0.29\linewidth}
    \footnotesize
    \captionof{figure}{\textbf{Learning across users and categories.}
    Varying personas, product categories, or both across episodes does not significantly change GPT-4o's performance.}
    \label{fig:user_experiments}
\end{minipage}

\end{figure}

Using GPT-4o with all other factors held constant, we evaluate whether agents learn differently when sequential experience varies along different axes.
Specifically, we compare settings where episodes vary by product category, by customer persona, or by both category and persona.
As shown in Figure~\ref{fig:user_experiments}, none of these variants produces a statistically significant difference in performance.
This suggests that GPT-4o does not reliably benefit from repeated experience, even when the structure of that experience is changed.

\subsection{Model Families}
\begin{figure}[H]
\centering

\begin{minipage}[c]{0.70\linewidth}
    \centering
    \includegraphics[width=\linewidth]{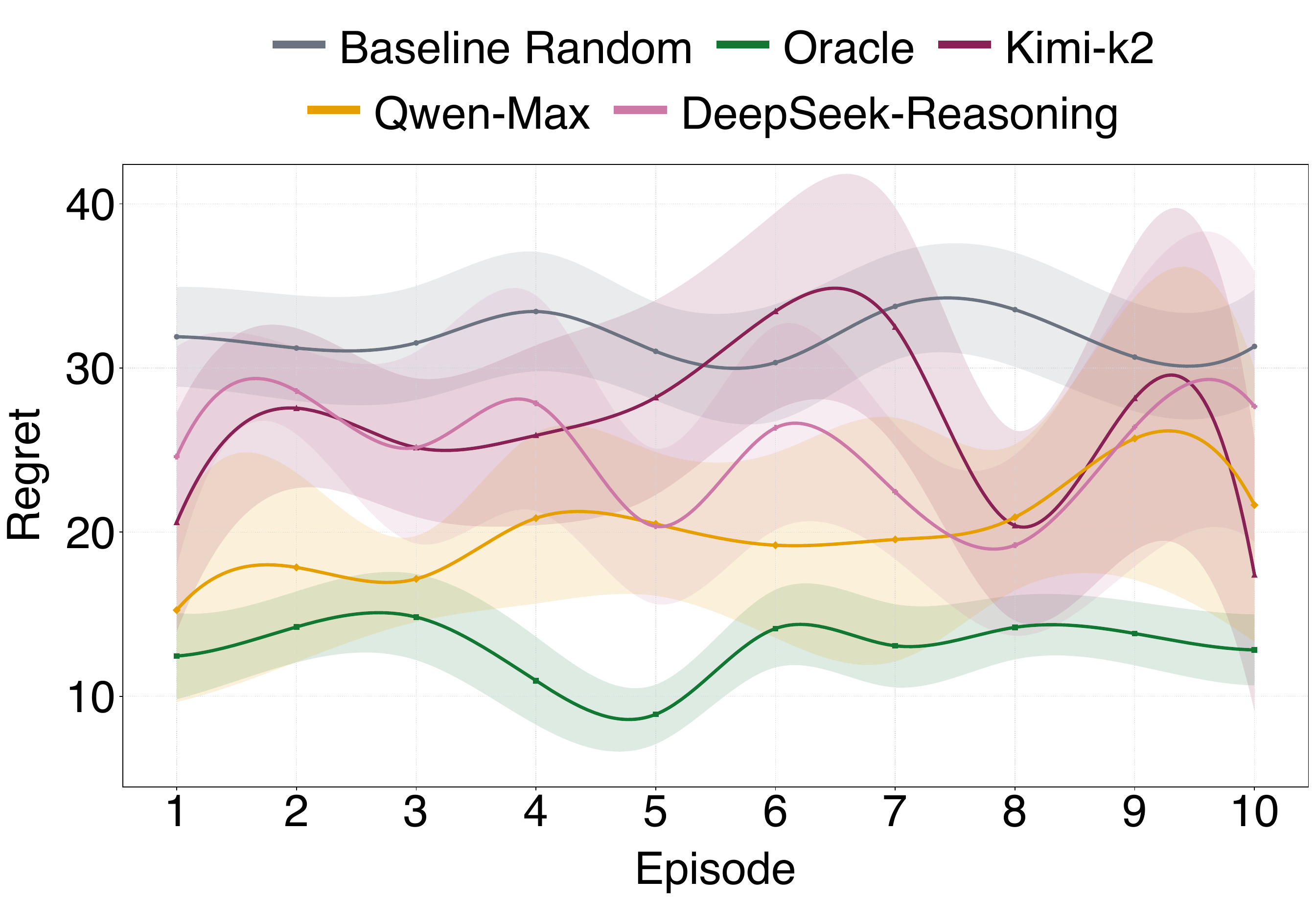}
\end{minipage}
\hfill
\begin{minipage}[c]{0.29\linewidth}
    \footnotesize
    \captionof{figure}{\textbf{Learning across model families.}
    Across GPT, Claude, Gemini, Kimi, Qwen, and DeepSeek model families, none of the evaluated models shows reliable learning across episodes.}
    \label{fig:model_families}
\end{minipage}

\end{figure}

We also evaluate whether the lack of experiential learning is specific to a small set of frontier models or persists across a broader model suite. We show additional results here for Kimi, Qwen, and DeepSeek. As shown in Figure~\ref{fig:model_families}, none of the evaluated models exhibits reliable episode-over-episode improvement. While models differ in absolute performance, their regret does not systematically decrease as they accumulate more interaction history. This suggests that the failure to learn from repeated customer experiences is not confined to a single model family, but is a broader limitation of current LLM-based recommendation agents.

\section{Model Confidence}
\label{app:ece}
As mentioned in the main text, we prompted the agent at every recommendation turn to report confidences for five binary targets: (1) chosen item is top-1 for the persona, (2) top-5, and (3–5) regret $\leq$ {5, 10, 20, 30}. For each target, we evaluated calibration per episode slice (Episodes 1, 5, and 10), pooling all turns across categories and seeds for that slice. We produced bin-wise accuracy-vs-confidence plots and summarized misalignment with a single score.
We ran the experiments with Baseline (no strategy) and DP-style planning prompts mentioned previously, and display the full grid of episode × target results for each condition.
\begin{figure}[H]
    \centering
    \includegraphics[width=\linewidth]{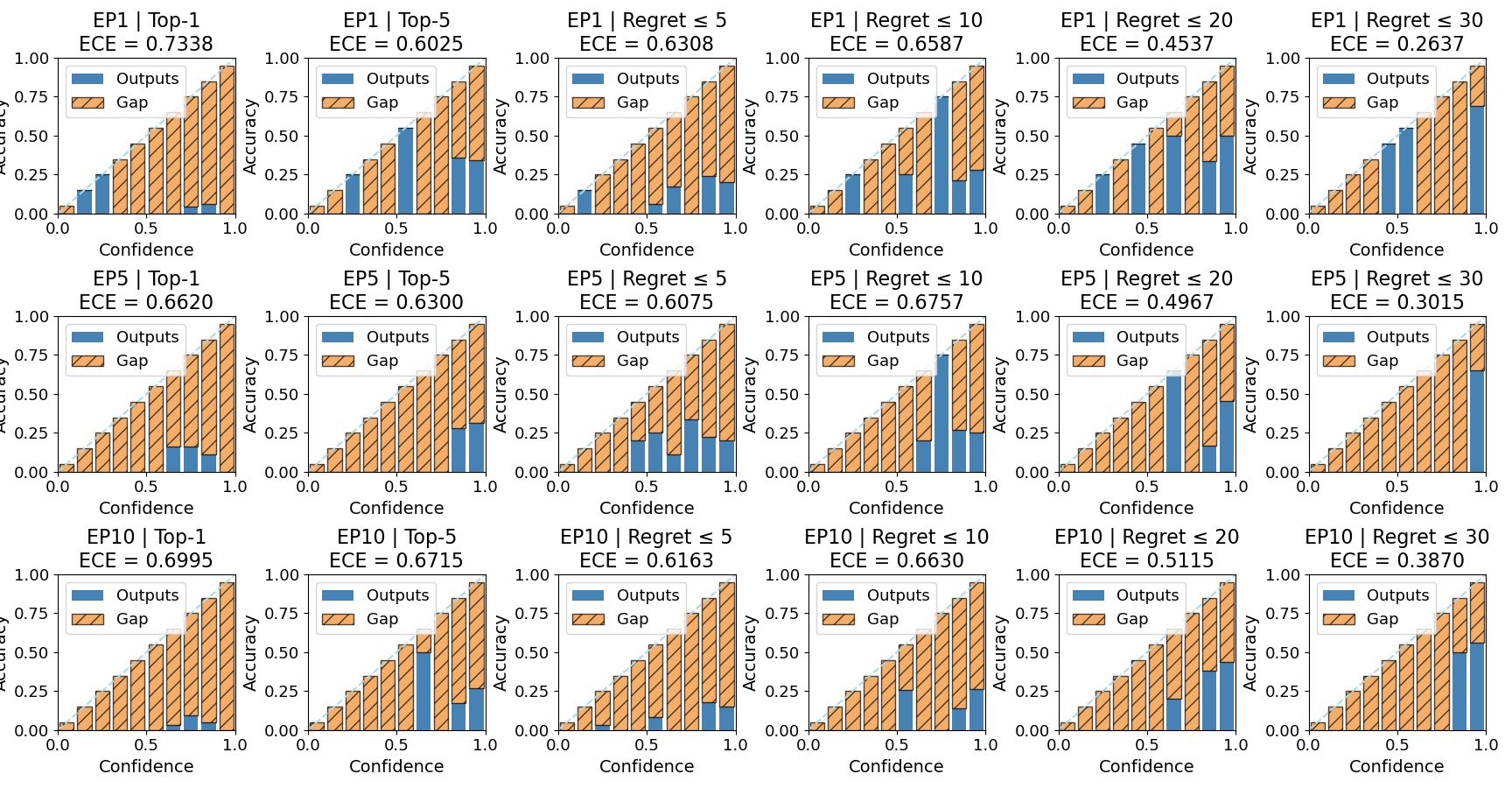}
    \caption{All ECE Runs using GPT-4o with No Strategy}
    \label{fig:ece_no_strat}
\end{figure}
\begin{figure}[H]
    \centering
    \includegraphics[width=\linewidth]{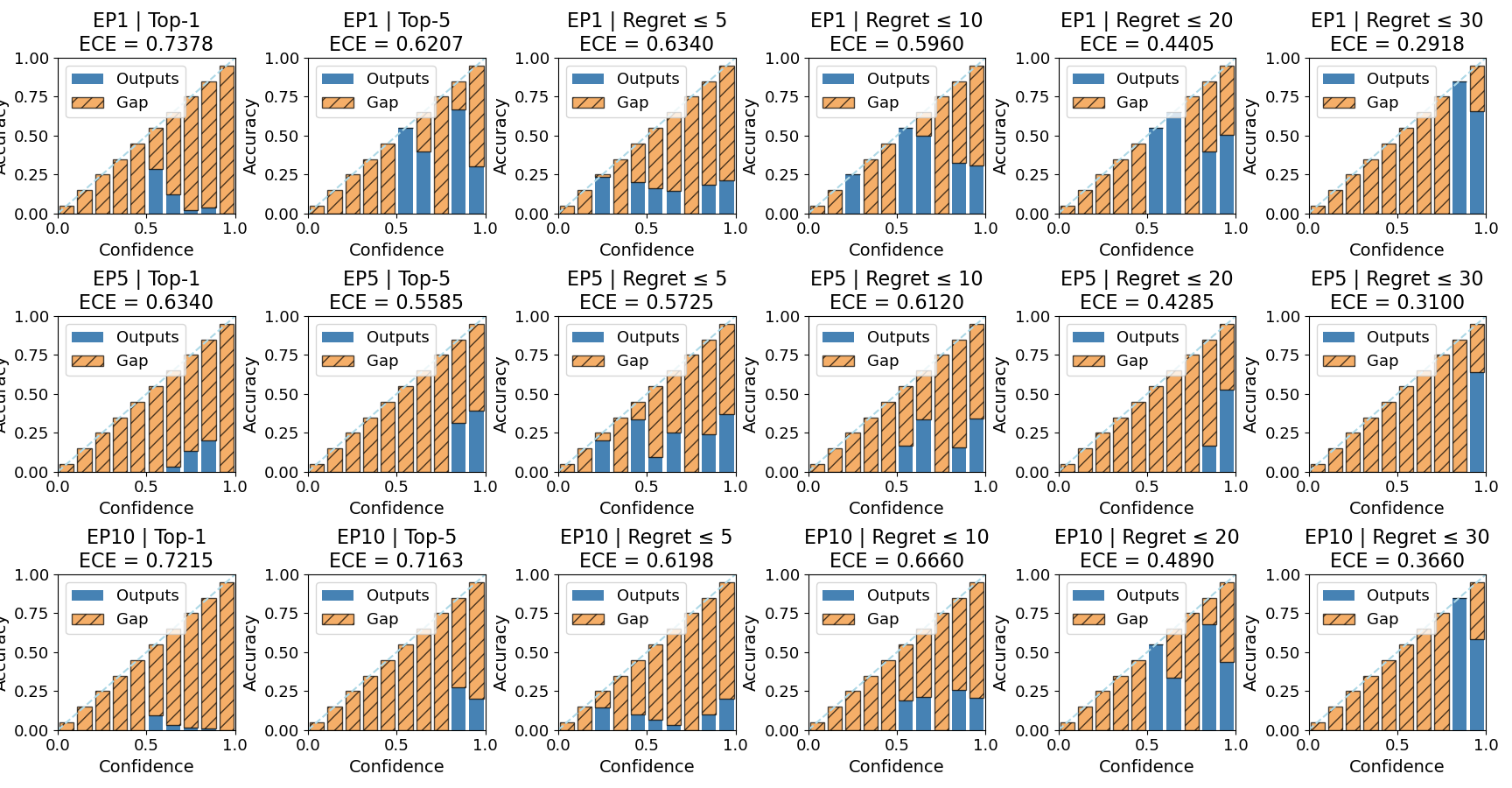}
    \caption{All ECE Runs using GPT-4o with DP-style Planning}
    \label{fig:ece_dp}
\end{figure}

\section{Manual Questioning Tests}
\label{app:manual}
To verify that poor agent performance is not due to an unsolvable environment, we introduce manual questioning experiments where a human designs the persona interactions. Figure~\ref{fig:manual_questions} shows an exemplar run in interactive mode, where it can be noted that carefully chosen questions enable recovery of the best product with zero regret.
\begin{figure}[H]
\begin{tcolorbox}[
  colback=lightblue!5!white,
  colframe=lightblue!100!white,
  title={An Example Manual Questions Experiment (Interactive Mode)},
  fonttitle=\bfseries,
  left=1.5mm,
  right=1.5mm,
  top=1mm,
  bottom=1mm
]

\small
\setlength{\parindent}{0pt}
\renewcommand{\arraystretch}{1.05}

\textbf{Running Manual Questions Experiment (Interactive Mode):}\\
Persona: 25905\\
Category: Office Racks \& Displays\\
Model: gpt-4o, Feedback: persona\\

\vspace{1mm}
\textbf{Category Office Racks \& Displays:} Max score 82.5, proceeding

\vspace{1mm}
\begin{tabularx}{\linewidth}{@{}>{\bfseries}p{5mm}@{\hspace{1mm}}X@{}}
Q: & How do you usually organize your desk at home or in the office? \\
A: & I organize my desk by keeping it tidy and functional, with my laptop in the center, a notepad and pen to the right, and any necessary documents or files neatly stacked on the left. \\[1mm]

Q: & Do you prefer everything stored neatly in trays, or do you like quick vertical access to files and folders? \\
A: & I prefer everything stored neatly in trays. \\[1mm]

Q: & How many papers, notebooks, or files do you typically keep on your desk at one time? \\
A: & Usually around three or four. \\[1mm]

Q: & Do your kids or family members often use your desk space, or is it mainly your own? \\
A: & It's mainly my own. \\[1mm]

Q: & Do you like darker neutral tones (black, brown, gray), or brighter finishes like gold or rose? \\
A: & I like darker neutral tones. \\[1mm]

Q: & Do you want a compact organizer to save space, or a larger one with multiple tiers for capacity? \\
A: & A compact organizer to save space. \\[1mm]

Q: & /done \\
\end{tabularx}

\vspace{1.5mm}
\textbf{Products in Office Racks \& Displays:} 26\\
\textbf{Top 3 products by score:}
\begin{enumerate}
    \setlength{\itemsep}{0pt}
    \setlength{\topsep}{1mm}
    \item SimpleHouseware Mesh Desk Organizer with Sliding D... (Score: 82.5)
    \item Figella Leather Office File Rack Document Filing T... (Score: 80.0)
    \item Febou 2 Pack Mail Organizer Desktop File Holder, 3... (Score: 80.0)
\end{enumerate}

\vspace{1mm}
\textbf{Final recommendation:} Product 107921\\
Score: 82.5, Best: 82.5\\
Top1: True, Top3: True\\
Feedback: \emph{Perfect! This is exactly what I was looking for. Great recommendation!}

\vspace{1mm}
Experiment completed successfully!\\
Chosen product rank: 1\\
Final regret: 0.0

\end{tcolorbox}
\caption{An exemplar run of the manual questions experiment in interactive mode.}
\label{fig:manual_questions}
\end{figure}

\section{Prompts}
\label{app:prompts}
\subsection{Oracle Agent}
We present the prompt used to define the oracle baseline. The oracle agent is given the full persona description and the complete set of candidate products, and is instructed to return only the index of the single best product. This setting establishes a lower bound on regret under full-information conditions.
\begin{figure}[H]
\begin{tcolorbox}[colback=orange!5!white,
                  colframe=orange!90!black,
                  title=Prompt for Oracle Agent Baseline,
                  fonttitle=\bfseries, left=1mm, right=1mm, top=1mm, bottom=1mm]
\small
\noindent
\textbf{System Message:} You are an oracle recommendation agent with perfect knowledge of a customer's preferences.

\medskip
\noindent
\textbf{Customer Persona:}\\
\texttt{\{self.persona\_description\}}

\medskip
\noindent
\textbf{Product Category:} \texttt{\{category\}}

\medskip
\noindent
\textbf{Available Products:}\\
\texttt{\{self.\_format\_products(products)\}}

\medskip
\noindent
\textbf{Task:} Given the customer's complete persona description, choose the single best product that would most satisfy their preferences and needs. You have perfect knowledge of what this customer would want.

\medskip
\noindent
\textbf{Output format (MUST be exactly one line, no extra text):}\\
\texttt{RECOMMEND: <array\_index\_0\_to\_\{num\_products-1\}>}

\medskip
\noindent
\textbf{Rules:}
\begin{itemize}
    \item Choose the product that best matches the customer's persona.
    \item Consider all aspects of their preferences, lifestyle, and needs.
    \item Return the array index (0-based), not the product ID.
    \item No explanations, just the recommendation index.
\end{itemize}
\end{tcolorbox}
\caption{Prompt used to instantiate the oracle baseline.}
\label{fig:prompt_oracle_baseline}
\end{figure}

\subsection{Persona Agents}
We introduce the prompts used for the persona agents. These agents are employed in three settings: (1) scoring personas, (2) generating dialogue, and (3) producing persona-based feedback.
\begin{figure}[H]
\begin{tcolorbox}[colback=orange!5!white,
                  colframe=orange!90!black,
                  title=Prompts for Customer Simulation,
                  fonttitle=\bfseries, left=1mm, right=1mm, top=1mm, bottom=1mm]
\footnotesize
\small

\begin{minipage}[t]{0.49\linewidth}
\textbf{(a) Preference Generation}\\[2pt]
\begin{tabularx}{\linewidth}{@{} l @{:\ } X @{}}
  ``\texttt{persona\_des}"  & ``..." \\
  ``\texttt{category}"    & ``..." \\
  ``\texttt{products}"    & ``[...]" \\
  ``\texttt{instructions}" & ``You ARE the persona described. Rate each product with a score from 0 to 100 (integers only) based on how much YOU would like it. Return a JSON object with key `results` as an array of objects: \{id, score\}. Do not include any other keys or text." \\
\end{tabularx}
\end{minipage}\hfill
\begin{minipage}[t]{0.49\linewidth}
\textbf{(b) Response Simulation}\\[2pt]
\begin{tabularx}{\linewidth}{@{} l @{:\ } X @{}}
  ``\texttt{persona\_des}"  & ``..." \\
  ``\texttt{question}"     & ``..." \\
  ``\texttt{instructions}" & ``You simulate a user with the given persona description. Answer \textbf{strictly} as this persona would: -- Only answer the question asked. -- Do not restate persona or add rationale. -- If a choice is requested, give one choice only. Return the answer as plain text." \\
\end{tabularx}
\end{minipage}

\end{tcolorbox}
\caption{Prompts for simulating customer behavior with persona agents: (a) generating product preference scores, (b) producing persona-consistent responses to agent queries.}
\label{fig:customer_prompts}
\end{figure}

\begin{figure}[H]
\begin{tcolorbox}[colback=orange!5!white,
                  colframe=orange!90!black,
                  title=Prompt for User Feedback Response,
                  fonttitle=\bfseries, left=1mm, right=1mm, top=1mm, bottom=1mm]
\small
\noindent
Reminder: You are a user with this persona:\\
\texttt{\{self.\_persona\_text\}}

\medskip
\noindent
A recommendation agent just suggested a product to you.

\medskip
\noindent
\textbf{Context:}
\begin{itemize}[leftmargin=2em]
\item \{\texttt{chosen\_info}\}
\item \{\texttt{conversation\_context}\}
\end{itemize}

\medskip
\noindent
\{\texttt{tone\_instruction}\}

\medskip
\noindent
\textbf{Task:} Respond naturally as this persona would—like you're talking to a helpful salesperson or friend. Be conversational and specific about your preferences. Keep it to 1–2 sentences and sound like a real person, not a formal review. Make it a statement about your preferences, not a question. Never mention specific scores, regret values, or reveal which product would be better.

\medskip
\noindent
\textbf{Your response:}
\end{tcolorbox}
\caption{Prompt for eliciting naturalistic persona feedback following a recommendation. The tone is adjusted based on the quality of the recommendation, estimated through regret.}
\label{fig:user_feedback_prompt}
\end{figure}

\subsection{Recommender Agent}
We describe the prompts used to instantiate the recommender agent. At its core, the agent is queried at each turn with a baseline prompt, shown in Figure~\ref{fig:agent_action_prompt}, which specifies the available context and requires the model to either ask one clarifying question or make a recommendation. \\
\\
To evaluate whether agents can exploit past interactions, we additionally replace the raw conversation history with a summary of prior episodes, producing the variant shown in Figure~\ref{fig:agent_summary_prompt}. \\
\\
To test whether explicit reasoning instructions improve performance, we append a chain-of-thought style enhancement to the baseline prompt, shown in Figure~\ref{fig:agent_cot_prompt}. These prompt designs define the controlled conditions under which we ablate model behavior, with results reported in Appendix~\ref{app:ablation}.
\begin{figure}[H]
\begin{tcolorbox}[colback=orange!5!white,
                  colframe=orange!90!black,
                  title=Prompt for Recommender Agent Action,
                  fonttitle=\bfseries, left=1mm, right=1mm, top=1mm, bottom=1mm]
\small
\noindent
\textbf{System Message:} You are a product recommendation agent. Your goal is to find the best product for this user, while asking the fewest number of questions before being confident in the best product for the user.

\medskip
\noindent
\textbf{Context:}\\
\texttt{\{context\}}  \\
\texttt{\{feedback\_context\}}

\medskip
\noindent
\textbf{Task:} First, analyze what you already know from the conversation. Then, either:
\begin{itemize}
\item Ask one short, consumer-friendly question to clarify user preferences, or
\item If sufficiently confident, recommend one product by index.
\end{itemize}

\noindent
\textbf{CRITICAL OUTPUT FORMAT (MUST FOLLOW EXACTLY):}
\begin{itemize}
\item To ask a question: \texttt{QUESTION: [your question here]}
\item To recommend: \texttt{RECOMMEND: [number between 0 and \{num\_products-1\}]}
\end{itemize}

\noindent
\textbf{STRICT RULES:}
\begin{itemize}
\item Your response must start with either ``\texttt{QUESTION:}'' or ``\texttt{RECOMMEND:}''.
\item Do NOT include any explanations, reasoning, or additional text.
\item Do NOT use bullets, multiple lines, or formatting.
\end{itemize}
\end{tcolorbox}
\caption{LLM prompt for the recommender agent to decide its next action (ask or recommend).}
\label{fig:agent_action_prompt}
\end{figure}

\begin{figure}[H]
\begin{tcolorbox}[colback=orange!5!white,
                  colframe=orange!90!black,
                  title=Prompts for Episode Generation Summary,
                  fonttitle=\bfseries, left=1mm, right=1mm, top=1mm, bottom=1mm]
\small
\noindent
\textbf{System Message:} You just completed Episode \texttt{\{episode\_num\}} in the \texttt{\{category\}} category for Persona \texttt{\{persona\}}.

\medskip
\noindent
\textbf{Episode Details:}\\
\texttt{\{dialog\_text\}}\\
Selected Product: \texttt{\{selected\_product\_id\}}\\
Feedback: \texttt{\{feedback\}}

\medskip
\noindent
\textbf{Your task:} Provide the context from this episode that you would want a future agent to know. Focus on:
\begin{itemize}
    \item What worked or didn't work in your approach.
    \item Key insights about user preferences or product selection.
    \item Any patterns you noticed that could help in similar situations.
\end{itemize}

\noindent
\textbf{Instruction:} Write only the summary, no additional commentary.
\end{tcolorbox}
\caption{LLM prompt variant for generating episode summaries.}
\label{fig:agent_summary_prompt}
\end{figure}
\begin{figure}[H]
\begin{tcolorbox}[colback=orange!5!white,
                  colframe=orange!90!black,
                  title=Prompts Enhancement for Chain-of-Thought (CoT),
                  fonttitle=\bfseries, left=1mm, right=1mm, top=1mm, bottom=1mm]
\small
\noindent
Let me think through this systematically:
\begin{itemize}
\item Customer preferences: [analyze what I know]
\item Available products: [analyze the options]
\item Best match: [reason about the best choice]
\item Decision: [decide whether to ask or recommend]
\end{itemize}

\noindent
\textbf{Let's reason step by step:}
\begin{enumerate}
\item What do I know about the customer so far?
\item What information am I still missing?
\item Based on this reasoning, what should I do next?
\end{enumerate}

\noindent
Before making your decision, think again: What are you unsure about regarding this customer? What questions should you ask next? Consider what additional information would help you make a better recommendation.

\medskip
\noindent
Think through each step carefully before responding.

\end{tcolorbox}
\caption{Chain-of-Thought (CoT) enhancement appended to agent prompts to test the reasoning process.}
\label{fig:agent_cot_prompt}
\end{figure}

\subsection{Planning Agents}
\label{app:planning_prompts}
In addition to the baseline recommender prompts described above, we introduce prompts designed to explicitly encourage planning behavior. These variants aim to test whether models can adopt more deliberate strategies for information gathering rather than defaulting to shallow heuristics.

The \textbf{Greedy} prompt (Figure~\ref{fig:prompt_greedy}) directs the agent to internally enumerate plausible candidate products and then select the single most informative clarifying question that would best differentiate among them.

The \textbf{POMDP} prompt (Figure~\ref{fig:prompt_dp_pomdp}) frames the interaction as a planning problem under uncertainty. Here, the agent is instructed to maintain a belief state over possible user preferences, evaluate the expected value of different candidate questions, and select the one with the highest information gain—even if its benefits only materialize in later turns.

Results are shown in Figure \ref{fig:planning}.

\begin{figure}[H]
\begin{tcolorbox}[colback=orange!5!white,
                  colframe=orange!90!black,
                  title=Prompt for Greedy Questioning,
                  fonttitle=\bfseries, left=1mm, right=1mm, top=1mm, bottom=1mm]
\small
\noindent
You are a product recommendation agent. Your goal is to find the best product for this user.

\medskip
\noindent
\textbf{Context:}\\
\texttt{\{context\}} \\
\texttt{\{feedback\_context\}}

\medskip
\noindent
\textbf{INTERNAL REASONING (do not share with customer):}
\begin{itemize}
\item First, list all the possible products that you think the customer might like based on what you know so far.
\item Then, think about what is the best question you could ask the customer to eliminate the most number of products from the list.
\end{itemize}

\medskip
\noindent
Your question should:
\begin{itemize}
\item Help you distinguish between the products you think the customer might like
\item Focus on the most important decision factor that's still unclear
\item Ask about preferences, needs, and requirements — NOT about specific products or product numbers
\end{itemize}

\medskip
\noindent
\textbf{CRITICAL OUTPUT FORMAT (MUST FOLLOW EXACTLY):}\\
\texttt{QUESTION: [your question here]}
\end{tcolorbox}
\caption{Greedy prompt used to bias the recommender agent toward asking the most informative single question at each turn.}
\label{fig:prompt_greedy}
\end{figure}
\begin{figure}[H]
\begin{tcolorbox}[colback=orange!5!white,
                  colframe=orange!90!black,
                  title=Prompts for Dynamic Programming (POMDP) Questioning,
                  fonttitle=\bfseries, left=1mm, right=1mm, top=1mm, bottom=1mm]
\small
\noindent
\textbf{System Message:} You are a product recommendation agent. Your goal is to find the best product for this user.

\medskip
\noindent
\textbf{Context:} \\
\texttt{\{context\}}  \\
\texttt{\{feedback\_context\}}

\medskip
\noindent
Think like a planner solving a \textbf{POMDP} with a \textbf{single terminal reward} from the score that the customer would assign to what you recommend. Note that you have \texttt{\{questions\_remaining\}} turn(s) left.

\begin{enumerate}
    \item \textbf{Maintain a belief state}—a probability distribution over possible customer preferences given past answers.
    \item For each possible next question:
        \begin{itemize}
            \item Predict how each possible answer will \textbf{update your belief}.
            \item Estimate how that updated belief will affect your \textbf{final recommendation quality}.
            \item Compute the \textbf{expected value of information (EVI)} for that question.
        \end{itemize}
    \item Choose the question with the \textbf{highest expected value}, even if it has no immediate payoff.
\end{enumerate}

\noindent
\textbf{Customer Interaction:} Ask the question that maximizes the expected value of information for your final recommendation.

\medskip
\noindent
\textbf{CRITICAL OUTPUT FORMAT (MUST FOLLOW EXACTLY):}\\
\texttt{QUESTION: [your question here]}

\end{tcolorbox}
\caption{POMDP-inspired prompt that instructs the agent to choose the question with the highest Expected Value of Information.}
\label{fig:prompt_dp_pomdp}
\end{figure}

\section{Additional Consistency Checks for the Customer Module}
\label{app:consistency}

A central requirement of our benchmark is that the customer simulator behaves \emph{consistently}: it should remain faithful to its persona specification and provide stable answers under repeated queries. We evaluate these properties through two complementary tests.

\subsection{Faithfulness to Persona Descriptions}

We first test whether simulator responses remain consistent with the underlying persona descriptions. For each of the 327 agent questions from 10 multi-episode trajectories, Claude Sonnet~4 labels the simulator's answer as \emph{Aligned}, \emph{Neutral}, or \emph{Conflict} with respect to the persona description. The raw breakdown is:
\[
\text{Aligned: } 244 \; (74.6\%), \qquad
\text{Neutral: } 63 \; (19.3\%), \qquad
\text{Conflict: } 20 \; (6.1\%).
\]
However, manual review shows that all 20 \emph{Conflict} labels are false positives. These cases typically arise when the judge infers contradictions from missing persona details, makes unwarranted assumptions about what activities a persona should enjoy, or treats differences in specificity as logical conflict. We identify no genuine contradictions between simulator responses and persona profiles.

\subsection{Response Stability Under Resampling}

We next measure whether the simulator gives semantically stable answers when asked the same question multiple times. Across 10 trajectories with 5 episodes each, we pause whenever the recommender asks a question and query the persona simulator independently 10 times, refreshing the GPT-4o persona agent for each run. We then present all 10 responses to Claude Sonnet~4, which identifies the majority position and labels each response as either aligned or misaligned with that position. Across 109 questions, this yields 1{,}090 total responses:
\[
\text{Aligned: } 1{,}029 \; (94.4\%), \qquad
\text{Misaligned: } 61 \; (5.6\%).
\]
Manual inspection indicates that most misaligned cases are not substantive preference changes, but minor numeric or stylistic variations. For example, in response to a budget question, answers such as ``under \$50,'' ``\$30 to \$50,'' ``moderately priced,'' and ``reasonably priced'' were sometimes separated by the evaluator despite expressing the same underlying value-conscious preference. Thus, the measured 94.4\% agreement rate should be interpreted as a conservative lower bound on semantic consistency.

\section{Dataset Collection}
\label{app:cat_matching}
\textbf{Collecting Products.} The Amazon Reviews dataset \citep{hou2024_amazonproducts}, consisting of 34 product categories, is set up such that one has to pick one of the 34 categories to sample products.
Therefore, to preserve the original distribution of categories, we performed weighted random sampling over the categories.
We provided the necessary scripts to draw more products in our codebase.

\paragraph{Matching Product Categories.}
The first challenge is inconsistent naming conventions; for example, a product's category path might be ...$\rightarrow$\texttt{Women} $\rightarrow$ \texttt{Shoes}, \texttt{Sandals}, whereas our standardized path is ... $\rightarrow$ \texttt{Women's Fashion} $\rightarrow$ \texttt{Women's Shoes} $\rightarrow$ \texttt{Women's Sandals}. The second challenge is structural discrepancy, where a product's path omits an intermediate level, such as \texttt{CDs\&Vinyl} $\rightarrow$ \texttt{Blues}, while the canonical path is \texttt{CDs\&Vinyl} $\rightarrow$ \texttt{Music Styles} $\rightarrow$ \texttt{Blues}. To solve these issues, our algorithm employs a two-pronged approach at each step of the path traversal:
\begin{itemize}
    \item Fuzzy Matching for Naming Inconsistencies: To resolve differing names, we apply a series of prioritized fuzzy matching heuristics. For instance, the substring detection heuristic is precisely what allows the algorithm to map a product's simpler category \texttt{Sandals} to the more descriptive canonical term \texttt{Women's Sandals}. Other heuristics, like word subset validation (mapping Action Figures to Action \& Toy Figures) and normalization (handling case and suffix differences), ensure robustness against a wide range of naming variations.
    \item Subtree Traversal for Structural Gaps: To handle missing intermediate levels, our algorithm does not merely search the immediate children of the last matched node. Instead, it performs a breadth-first search through the entire subtree of descendants. This is how it bridges structural gaps. In the example \texttt{CDs \& Vinyl} $\rightarrow$ \texttt{Blues}, once the algorithm successfully matches the \texttt{CDs \& Vinyl} node, it then searches all descendants for a \texttt{Blues} node. It will find the correct Blues node even though it is a grandchild (nested under \texttt{Music Styles}), effectively ``skipping over" the missing level in the product's path data.
\end{itemize}
This matching enables us to successfully map about 70\% of the sampled products to a valid path within the target hierarchy. Despite the flexibility of these heuristics, the overall assignment criteria remain strict: a product is considered successfully matched only when its entire category sequence corresponds to a valid, continuous path from the root.

\paragraph{Final Database.}
From the remaining category levels, we extracted individual category levels and flattened the hierarchical structure into a relationship database schema. The processed data was organized into three tables: (1) a products table containing product metadata, (2) a categories table, and (3) a product-category table implementing a many-to-many relationship between products and categories. From this database, we then pruned categories that were still too broad (e.g., Cooking \& Baking) and would not be searched for by the typical user and niche categories that had fewer than 15 products.
The final database consists of 71088 products, 2030 categories, and 100485 product-category links.


\section{Additional Models' Failure Modes}
In this section, we discuss some notable failure modes of the considered models.

\paragraph{Repetitive Questions.}
A recurring issue with some models was an endless loop of repetitive questions. At times, the model would repeatedly ask the same queries despite having access to the full conversation history, continuing until reaching a 20-turn limit. This was particularly evident with GPT-4o in all three experiments. Both Gemini models did not have this issue, while both Claude models showed repeated questions only under the variable setting experiment mode.

\paragraph{Failure to Memorize Requirements.}
Occasionally, explicit preferences stated by the customer are ignored.
The final product chosen by the recommender agent can go against the stated preferences directly.
This shows that SOTA models still struggle to respect specifications provided in context.

\paragraph{Model Specific Questions.}
Models from the Gemini series exhibit their own specific rigid patterns. They frequently ask questions like,``Are you looking for a free option?'' and then often proceed immediately to a recommendation. This behavior suggests the model is following a pre-determined conversational script rather than dynamically adapting to the user's needs, showing no interest in asking more questions and prematurely ending the conversation.

\section{Model Use And Cost Analysis}
\input{tables/model_use}

Table~\ref{tab:model_usage} details the model used in each of the experiment. 
One run spanning 10 episodes costs roughly 5 dollars between the recommender and the customer simulator.
Overall, reproducing all of the experiments in this work would cost $\sim$4,000 dollars.

%% file: tables/model_use.tex
\begin{table*}[t]
\centering
\small
\renewcommand{\arraystretch}{1.18}
\setlength{\tabcolsep}{5pt}

\begin{tabular}{
@{}
>{\raggedright\arraybackslash}p{0.20\linewidth}
>{\raggedright\arraybackslash}p{0.21\linewidth}
>{\raggedright\arraybackslash}p{0.40\linewidth}
>{\raggedright\arraybackslash}p{0.15\linewidth}
@{}}
\toprule
\textbf{Experiment} & \textbf{Model(s)} & \textbf{Key setting} & \textbf{Reported in} \\
\midrule

\textbf{Main recommendation experiments}
&
GPT-5.4; Gemini-3.1-Pro-Preview; Claude-Opus-4.6
&
Setting (a); 10 episodes; 40 random seeds; Free-form Text feedback
&
Fig.~\ref{fig:results1} left; Fig.~\ref{fig:turn_vs_episode}
\\
\midrule

\textbf{Full-information oracle baseline}
&
Claude-Sonnet-4
&
Full persona description provided; directly selects a product
&
Fig.~\ref{fig:results1} left; Fig.~\ref{fig:turn_vs_episode}
\\
\midrule

\textbf{Simple baselines}
&
Random; Popularity
&
Random: uniform product sampling; Popularity: highest-rated product, tie-broken by lower price
&
Fig.~\ref{fig:results1} left; Appendix
\\
\midrule

\textbf{Confidence / uncertainty evaluation}
&
GPT-4o
&
Setting (a); 10 episodes; turn-level recommendations in Episodes 1, 5, and 10
&
Fig.~\ref{fig:results1} right; Appendix~\ref{app:ece}
\\
\midrule

\textbf{Ideal trajectory construction}
&
GPT-5.4
&
5 candidate trajectories per episode; one selected and frozen as context; used for later episodes
&
Fig.~\ref{fig:turn_vs_episode}; Fig.~\ref{fig:ideal_traj}
\\
\midrule

\textbf{Additional ablations}
&
Kimi-K2; Qwen-Max; DeepSeek-Reasoning; Gemini-2.5-Pro; Gemini-2.5-Flash
&
Additional settings, feedback types, prompts, and question-count analysis for Gemini models
&
Appendix~\ref{app:ablation}
\\

\bottomrule
\end{tabular}
\caption{Model usage and experimental settings.}
\label{tab:model_usage}

\end{table*}